\documentclass[10pt,twocolumn,letterpaper]{article}
\usepackage[table,dvipsnames]{xcolor}
\newif\ifarxiv
\arxivtrue

\ifarxiv    \usepackage[pagenumbers]{iccv} 
\else       \usepackage{iccv}              
\fi

\definecolor{iccvblue}{rgb}{0.21,0.49,0.74}
\usepackage[pagebackref,breaklinks,colorlinks,allcolors=iccvblue]{hyperref}
\usepackage{algorithm}
\usepackage{algpseudocode}
\usepackage[rightComments=false]{algpseudocodex}
\usepackage{amsmath}

\usepackage{graphbox}
\usepackage{multirow}
\usepackage{makecell}
\usepackage{wrapfig}
\usepackage{epsfig}
\usepackage{pifont}

\usepackage{amsmath,amsfonts,bm,amssymb,mathtools,amsthm}
\usepackage{color,xcolor,xspace}
\usepackage{booktabs}
\usepackage{thm-restate}

\def\eqref#1{Eq.~(\ref{#1})}

\def\1{\bm{1}}

\def\vx{{\bm{x}}}

\DeclareMathAlphabet{\mathsfit}{\encodingdefault}{\sfdefault}{m}{sl}
\SetMathAlphabet{\mathsfit}{bold}{\encodingdefault}{\sfdefault}{bx}{n}

\DeclareMathOperator*{\argmin}{arg\,min}

\newcommand{\tabVsSmoothquant}{%
\begin{table}%
\centering
\scriptsize
\setlength{\tabcolsep}{6pt} 
\begin{tabular}{lcrr}
    \toprule
    \textbf{Method} & \textbf{Weight Quant. Error} & \textbf{FID}$\downarrow$ & \textbf{sFID}$\downarrow$  \\
    \midrule
    Baseline \cite{huang2024tfmq} & 0.0060 & 36.08 & 33.06 \\
    Baseline + SmoothQuant \cite{xiao2023smoothquant} & 0.0694 & 454.16 & 254.41 \\
    \rowcolor{gray!20} LES (ours) & \textbf{0.0058} & \textbf{30.37} & \textbf{22.72} \\
    \bottomrule
\end{tabular}
\vspace{-3mm}
\caption{Comparison on FFHQ $256\times256$ in W4A8.}
\vspace{-5mm}
\label{tab:vs_smoothquant}
\end{table}%
}%

\newcommand{\tabUncond}{%
\begin{table*}{%
\centering
\small
\begin{tabular}{lccccccccc}
    \toprule
    \multirow{2}{*}{\textbf{Method}} & \multirow{2}{*}{\textbf{Bits}} & \multicolumn{2}{p{0.23\textwidth}}{\centering \textbf{FFHQ (LDM-4)}} & \multicolumn{2}{p{0.23\textwidth}}{\centering \textbf{LSUN-Bedroom (LDM-4)}} & \multicolumn{2}{p{0.23\textwidth}}{\centering \textbf{LSUN-Church (LDM-8)}}\\ \cmidrule(r){3-4} \cmidrule(r){5-6} \cmidrule(r){7-8}
    & (W/A) & FID$\downarrow$ & sFID$\downarrow$ & FID$\downarrow$ & sFID$\downarrow$ & FID$\downarrow$ & sFID$\downarrow$\\
    \midrule
    Full Prec. & 32/32 & 31.34 & 25.88 & 9.82  & 20.17  & 11.28 & 29.58  \\
    \midrule
    Q-Diffusion~\cite{li2023q} & 8/8 & 33.10 & 27.69 & 14.79 & 27.36 & 12.87 & 30.68  \\
    PTQD~\cite{he2024ptqd} & 8/8 & 31.87 & \underline{25.91} & 12.90 & \underline{18.56} & 12.73 & 30.00 \\
    EDA-DM~\cite{liu2024enhanced} & 8/8 & \underline{31.28} & 31.25 & \underline{9.87} & 20.29 & 12.31 & 29.82 \\
    TFMQ-DM~\cite{huang2024tfmq} & 8/8 & 32.35 & 26.78 & 10.05 & 21.12 & \underline{11.25} & \textbf{29.18} \\
    \rowcolor{gray!20} Ours & 8/8 & \textbf{26.78} & \textbf{20.77} & \textbf{8.52} & \textbf{14.33} & \textbf{11.16} & \underline{29.35}\\
    \midrule
    
    Q-Diffusion~\cite{li2023q} & 4/8 & 36.17 & \underline{28.75} & 14.87 & 27.60 & 15.42 & 33.02 \\
    PTQD~\cite{he2024ptqd} & 4/8 & 36.65 & 32.52 & 19.27 & 20.59 & 14.52 & 33.03 \\
    EDA-DM~\cite{liu2024enhanced} & 4/8 & 36.28 & 31.07 & 16.96 & 31.64 & \textbf{11.40} & \textbf{28.80} \\
    TFMQ-DM~\cite{huang2024tfmq} & 4/8 & \underline{36.08} & 33.06 & \underline{10.65} & \underline{18.99} & 14.28 & 34.62 \\
     \rowcolor{gray!20} Ours & 4/8 & \textbf{30.37} & \textbf{22.72} & \textbf{9.73} & \textbf{15.94} & \underline{11.58} &  \underline{30.87} \\
    \midrule
    
    Q-Diffusion~\cite{li2023q} & 4/6 & 71.16 & 75.70 & 64.31 & 35.09 & 57.34 & 58.58 \\
    PTQD~\cite{he2024ptqd} & 4/6 & 74.18 & 80.51 & 106.21 & 39.44 & 56.90 & 57.19 \\
    EDA-DM~\cite{liu2024enhanced} & 4/6 & \underline{29.62} & 28.59 & 75.21 & 49.18 & \underline{14.41} & \underline{32.79} \\
    TFMQ-DM~\cite{huang2024tfmq} & 4/6 & 29.76 & \underline{27.07} & \underline{17.03} & \underline{34.27} & 26.74 & 57.51 \\
    \rowcolor{gray!20} Ours & 4/6 & \textbf{26.38} & \textbf{20.01} & \textbf{11.26} & \textbf{16.97} & \textbf{14.28} &  \textbf{32.76} \\
    \bottomrule
\end{tabular}
}
\vspace{-2mm}
\caption{
Quantization results of unconditional generation across various datasets at 256$\times$256 resolution.
Values in \textbf{boldface} indicate the best results, while \underline{underlined} values indicate the second-best. W/A indicates the bit-width of the weights and activations, respectively.}
\label{tab:uncond}
\vspace{-4mm}
\end{table*}%
}%

\newcommand{\tabCond}{%
\begin{table}%
\centering
\footnotesize 
\setlength{\tabcolsep}{2.5pt} 
\begin{tabular}{l@{\hspace{-3pt}}ccccccc}
    \toprule
    \multirow{2}{*}{\textbf{Method}} & \multirow{2}{*}{\textbf{Bits}} & \multicolumn{6}{c}{\textbf{ImageNet (LDM-4)}} \\ 
    \cmidrule(lr){3-8}
    & (W/A) & IS$\uparrow$ & FID$\downarrow$ & sFID$\downarrow$ & LPIPS$\downarrow$ & SSIM$\uparrow$ & PSNR$\uparrow$ \\
    \midrule
    Full Precision & 32/32 & 366.8 & 11.34 & 7.81 & -- & -- & -- \\
    \midrule
    Q-Diffusion~\cite{li2023q} & 8/8 & 351.3 & 11.11 & 12.20 & 0.116 & 0.895 & 26.88 \\
    PTQD$^\dagger$~\cite{he2024ptqd} & 8/8 & 349.6 & 10.94 & 11.28 & -- & -- & -- \\
    EDA-DM~\cite{liu2024enhanced} & 8/8 & 303.2 & 11.85 & 23.00 & 0.385 & 0.722 & 19.19 \\
    TFMQ-DM~\cite{huang2024tfmq} & 8/8 & \textbf{362.7} & 11.17 & 8.19 & 0.072 & 0.933 & 29.18 \\
    \rowcolor{gray!20} Ours & 8/8 & 357.9 & \textbf{10.29} & \textbf{7.38} & \textbf{0.058} & \textbf{0.945} & \textbf{31.14} \\
    \midrule
    Q-Diffusion~\cite{li2023q} & 4/8 & 330.4 & 9.34 & 12.39 & 0.170 & 0.846 & 23.97 \\
    PTQD$^\dagger$~\cite{he2024ptqd} & 4/8 & 321.7 & 9.04 & 13.09 & -- & -- & -- \\
    EDA-DM~\cite{liu2024enhanced} & 4/8 & 312.9 & 9.86 & 14.22 & 0.237 & 0.803 & 22.37 \\
    TFMQ-DM~\cite{huang2024tfmq} & 4/8 & 342.1 & \textbf{9.51} & 8.10 & 0.181 & 0.848 & 23.30 \\
    \rowcolor{gray!20} Ours & 4/8 & \textbf{350.8} & 9.68 & \textbf{7.19} & \textbf{0.124} & \textbf{0.886} & \textbf{25.90} \\
    \midrule
    Q-Diffusion~\cite{li2023q} & 4/6 & 65.2 & 36.49 & 71.41 & 0.477 & 0.613 & 17.09 \\
    PTQD$^\dagger$~\cite{he2024ptqd} & 4/6 & 73.1 & 33.73 & 60.80 & -- & -- & -- \\
    EDA-DM~\cite{liu2024enhanced} & 4/6 & 46.8 & 48.23 & 106.05 & 0.573 & 0.572 & 15.91 \\
    TFMQ-DM~\cite{huang2024tfmq} & 4/6 & 225.6 & 9.61 & 10.19 & 0.336 & 0.730 & 19.83 \\
    \rowcolor{gray!20} Ours & 4/6 & \textbf{320.6} & \textbf{7.81} & \textbf{7.26} & \textbf{0.194} & \textbf{0.845} & \textbf{23.70} \\
    \bottomrule
\end{tabular}
\vspace{-2mm}
\caption{Quantization results of class conditional generation. $\dagger$: PTQD utilizes noise correction during the sampling, making it unable to compare pixel level similarities even under the same seed. }
\label{tab:cond}
\end{table}%
}%

\newcommand{\tabTxtToimg}{%
\begin{table}
\centering
\scriptsize
\setlength{\tabcolsep}{2pt} 
\begin{tabular}{l@{\hspace{-3pt}}cccccccc}
    \toprule
    \multirow{2}{*}{\textbf{Methods}} & \multirow{2}{*}{\textbf{Bits}} & \multicolumn{7}{c}{\textbf{MS-COCO (Stable Diffusion)}} \\ 
    \cmidrule(lr){3-9}
    & (W/A) & \textbf{IS$\uparrow$} & \textbf{FID$\downarrow$} & \textbf{sFID$\downarrow$} & \textbf{CLIP$\uparrow$} & \textbf{LPIPS$\downarrow$} & \textbf{SSIM$\uparrow$} & \textbf{PSNR$\uparrow$} \\
    \midrule
    Full Precision & 32/32 & 40.23 & 20.38 & 62.55 & 31.27 & -- & -- & -- \\
    \midrule
    EDA-DM~\cite{liu2024enhanced} & 4/8 & 37.78 & 20.61 & 63.67 & 31.10 & 0.460 & 0.518 & 14.33 \\
    TFMQ-DM~\cite{huang2024tfmq} & 4/8 & 37.58 & 19.36 & \textbf{60.94} & 31.18 & 0.468 & 0.54 & 15.12 \\
    \rowcolor{gray!20} Ours & 4/8 & \textbf{40.16} & \textbf{19.20} & 60.96 & \textbf{31.41} & \textbf{0.443} & \textbf{0.557} & \textbf{15.34} \\
    \midrule
    EDA-DM~\cite{liu2024enhanced} & 4/6 & 27.26 & 33.23 & 67.06 & 29.73 & 0.623 & 0.360 & 12.34 \\
    TFMQ-DM~\cite{huang2024tfmq} & 4/6 & 16.22 & 104.95 & 126.13 & 25.32 & 0.691 & 0.361 & 12.69 \\
    \rowcolor{gray!20} Ours & 4/6 & \textbf{33.10} & \textbf{29.58} & \textbf{67.05} & \textbf{30.67} & \textbf{0.537} & \textbf{0.427} & \textbf{14.05} \\
    \bottomrule
\vspace{-4mm}
\end{tabular}
\caption{Quantization results of text-guided image generation. }
\vspace{-1mm}
\label{tab:t2i}
\end{table}
}%

\newcommand{\tabAblation}{%
\begin{table}%
\centering
\footnotesize
\setlength{\tabcolsep}{3.5pt} 
\begin{tabular}{lcccccc}
    \toprule
    \textbf{Method} & \textbf{Bits (W/A)} & \textbf{IS}$\uparrow$ & \textbf{FID}$\downarrow$ & \textbf{sFID}$\downarrow$  \\
    \midrule
    Full Precision & 32/32 & 3.70 & 31.34 & 25.88 &  \\
    \midrule
    Baseline & 4/8 & 3.50 & 36.08 & 33.06 &  \\
    + Learned Equivalent Scaling & 4/8 & 3.60 & 33.46 &26.29 &  \\
    + Adaptive Timestep Weighting & 4/8 & 3.69 & 31.83 & 24.39 &  \\
    \rowcolor{gray!20} + Power-of-Two Scaling & 4/8 & \textbf{3.75} & \textbf{30.37} & \textbf{22.72} &  \\
    \bottomrule
\end{tabular}
\vspace{-2mm}
\caption{Ablation study on components of the proposed method.}
\vspace{-2mm}
\label{tab:ablation}
\end{table}%
}%

\newcommand{\tabAblationWeighting}{%
\begin{table}%
\centering
\footnotesize
\setlength{\tabcolsep}{4pt} 
\begin{tabular}{lcccccc}
    \toprule
    \textbf{Method} & \textbf{Bits (W/A)} & \textbf{IS}$\uparrow$ & \textbf{FID}$\downarrow$ & \textbf{sFID}$\downarrow$  \\
    \midrule
    Full Precision & 32/32 & 3.70 & 31.34 & 25.88 &  \\
    \midrule
    Uniform & 4/8 & 3.60 & 33.46 &26.29 &  \\
    Linear & 4/8 & 3.59 & 36.58 & 30.20 &  \\
    Quadratic & 4/8 & 3.61 & 35.30 & 27.39 &  \\
    \rowcolor{gray!20} Adaptive Timestep Weighting & 4/8 & \textbf{3.69} & \textbf{31.83} & \textbf{24.39} &  \\
    \bottomrule
\end{tabular}
\vspace{-2mm}
\caption{Ablation study on timestep weighting strategy.}
\vspace{-2mm}
\label{tab:ablation_weighting}
\end{table}%
}%

\newcommand{\tabAblationPot}{%
\begin{table}%
\centering
\footnotesize
\setlength{\tabcolsep}{4pt} 
\begin{tabular}{lccccccc}
    \toprule
    \textbf{Method} & \textbf{Target layers} & \textbf{Bits (W/A)} & \textbf{IS}$\uparrow$ & \textbf{FID}$\downarrow$ & \textbf{sFID}$\downarrow$  \\
    \midrule
    Full Precision & & 32/32 & 3.70 & 31.34 & 25.88 &  \\
    \midrule
    w/o PTS & None & 4/8 & 3.69 & 31.83 & 24.39 \\
    MSE & all layers & 4/8 & 3.74 & 32.55 & 24.24 &  \\
    Voting & all layers & 4/8 & 3.72 & 31.91 & 23.86 &  \\
    \rowcolor{gray!20} Voting & skip connect. & 4/8 & \textbf{3.75} & \textbf{30.37} & \textbf{22.72} &  \\
    \bottomrule
\end{tabular}
\vspace{-2mm}
\caption{Ablation study on PTS scaling.}
\vspace{-2mm}
\label{tab:ablation_pot}
\end{table}%
}%

\newcommand{\tabHyper}{%
\begin{table*}[ht!]
    \centering
    \begin{tabular}{lcccccccccc}
    \toprule
    Experiment  & $T$ & $n$ & $cfg$ & $N$ & $\alpha$ & $\kappa$ & $\xi$ & $D$ & $B$ & $iteration$ \\ 
    \midrule
    LDM-8 LSUN-Church (uncond.) & 20 & 256 & -- & 5120  & 25 & 0.85 & 0.95 & 3 & 32 & 4000 \\
    LDM-4 LSUN-Bedroom (uncond.) & 20 & 256 & -- & 5120  & 20 & 0.85 & 0.95 & 3 & 32 & 6000 \\
    LDM-4 FFHQ (uncond.) & 20 & 256 & -- & 5120  & 20 & 0.85 & 0.95 & 3 & 32 & 6000 \\
    LDM-4 ImageNet (class cond.) &  20 & 256 & \ding{51} & 10240  & 20 & 0.85 & 0.95 & 3 & 32 & 6000 \\
    Stable Diffusion (text cond.) & 25 & 256 & \ding{55} & 6400  & 20 & 0.85 & 0.95 & 3 & 8 & 6000 \\
    \bottomrule
    \end{tabular}
    \caption{Hyperparameters for all experiments. $T/c$ is the number of sampling steps for generating the calibration data. $n$ is the amount of calibration data per sampling step, and $N$ is the size of calibration dataset. \emph{cfg} indicates whether classifier-free guidance was used or not. $B$ is the batch size. \emph{iteration} is the number of training steps used to learn $\tau$. }
    \label{tab:hyper}
\end{table*}
}
\newcommand{\figFirstPage}{
  \vspace{-5mm}
	\centering
	\includegraphics[width=\textwidth, trim=0em 0em 0em 0em, clip]{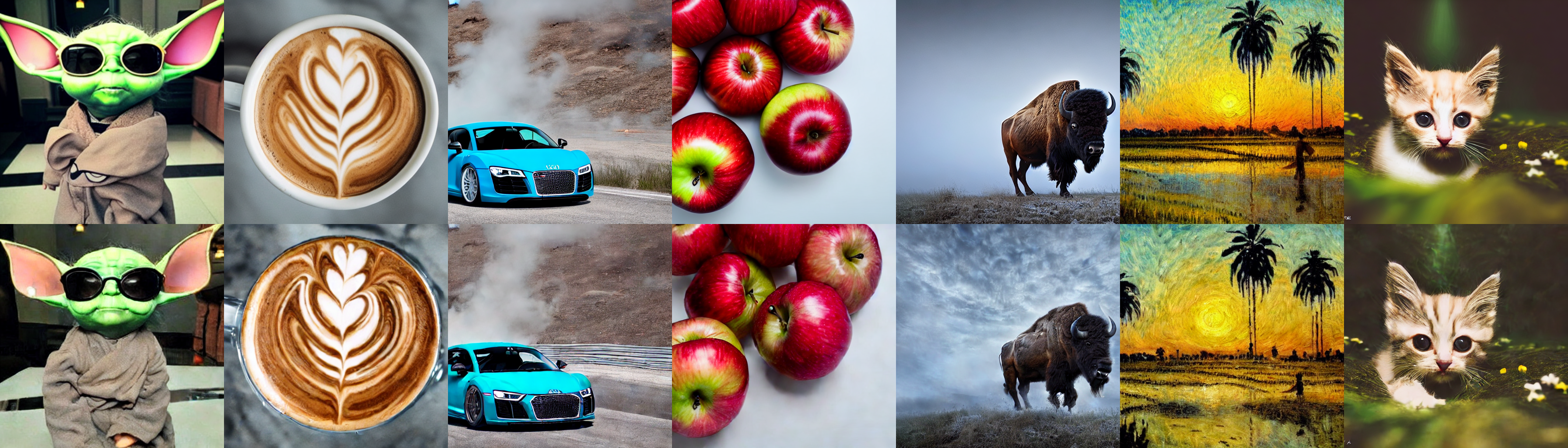}
  \vspace{-5mm}
  \captionof{figure}{
  Images generated by Stable Diffusion~\cite{rombach2022high} in full precision (top row) and in W4A8-quantization using our method (bottom row).
  }
	\label{fig:frontpage}
  \vspace{3.5mm}
}

\newcommand{\figBeforeAfter}{
\begin{figure*}[!t]
    \centering
    \begin{subfigure}[t]{0.19\linewidth}
        \centering
        \includegraphics[width=\textwidth]{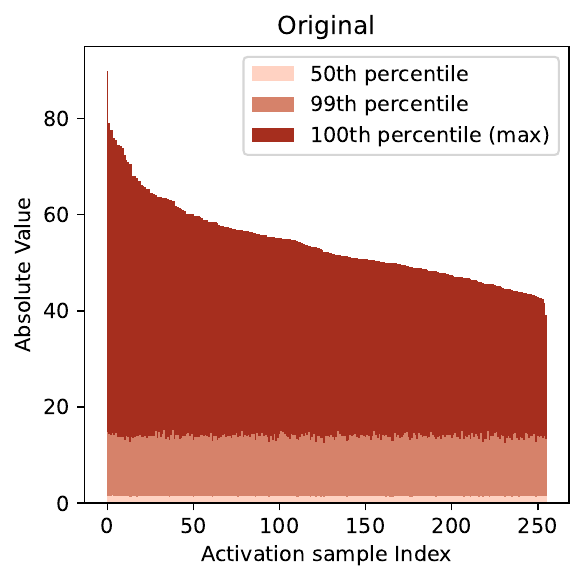}
        \subcaption{$\mathbf{X}$}
    \end{subfigure}
    \begin{subfigure}[t]{0.19\linewidth}
        \centering
        \includegraphics[width=\textwidth]{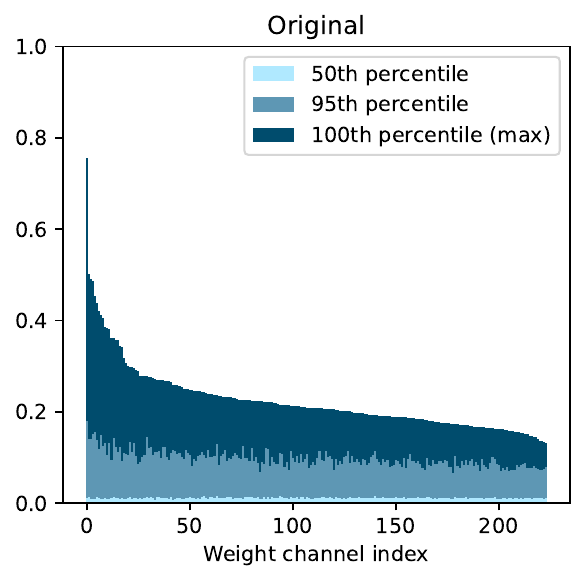}
        \subcaption{$\mathbf{W}$}
    \end{subfigure}
    \begin{subfigure}[t]{0.003\linewidth}
        \centering
        \includegraphics[width=\textwidth]{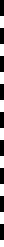}
    \end{subfigure}
    \begin{subfigure}[t]{0.19\linewidth}
        \centering
        \includegraphics[width=\textwidth]{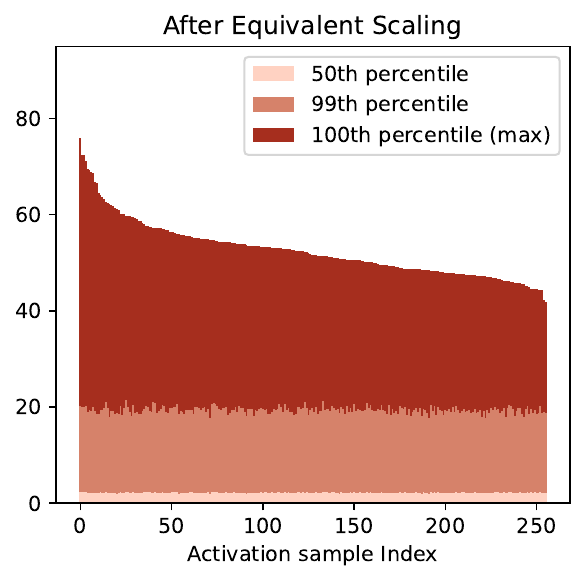}
        \subcaption{$\mathbf{\hat{X}} = [\mathbf{X} \oslash \tau]$}
    \end{subfigure}
    \begin{subfigure}[t]{0.19\linewidth}
        \centering
        \includegraphics[width=\textwidth]{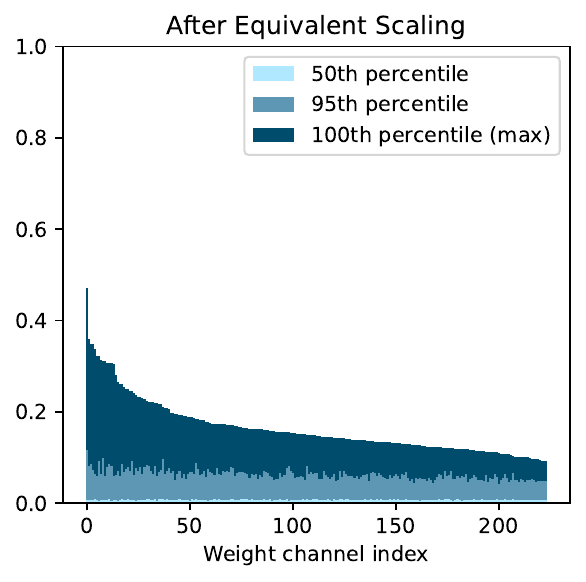}
        \subcaption{$\mathbf{\hat{W}} = [\tau^\top \odot \mathbf{W}]$}
    \end{subfigure}
    \begin{subfigure}[t]{0.003\linewidth}
        \centering
        \includegraphics[width=\textwidth]{res/line.png}
    \end{subfigure}
    \begin{subfigure}[t]{0.19\linewidth}
        \centering
        \includegraphics[width=\textwidth]{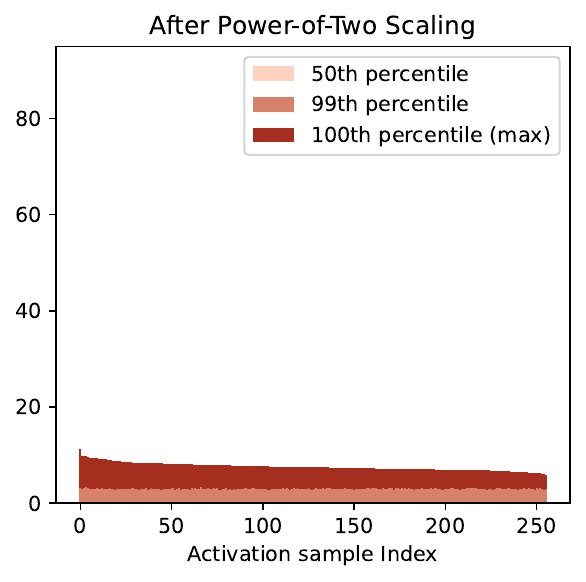}
        \subcaption{$\mathbf{\hat{X}} \odot \left(1/2^{\boldsymbol{\delta}}\right)$}
    \end{subfigure}
     \vspace{-2mm}
    \caption{Visualization of activations and weights in a skip connection layer, before and after applying Equivalent Scaling and Power-of-Two Scaling. Each values are sorted along x-axis by their maximum values. Darker colors indicate higher values, representing these prominent outliers. In the original distributions (a) and (b), both $\mathbf{X}$, $\mathbf{W}$ contain large outliers.
    After applying Equivalent scaling, large outliers are reduced in both the scaled activation $\mathbf{\hat{X}}$ and weights $\mathbf{\hat{X}}$, as shown in (c) and (d). When Power-of-Two Scaling is applied as in (e), the activation outliers are further diminished, demonstrating the effectiveness of PoT scaling in reducing extreme activation values.}
     \vspace{-5mm}
     \label{fig:before_after}
\end{figure*}
}

\newcommand{\figFidMse}{%
\begin{figure}[t]
\begin{center}
\includegraphics[width=1\columnwidth]{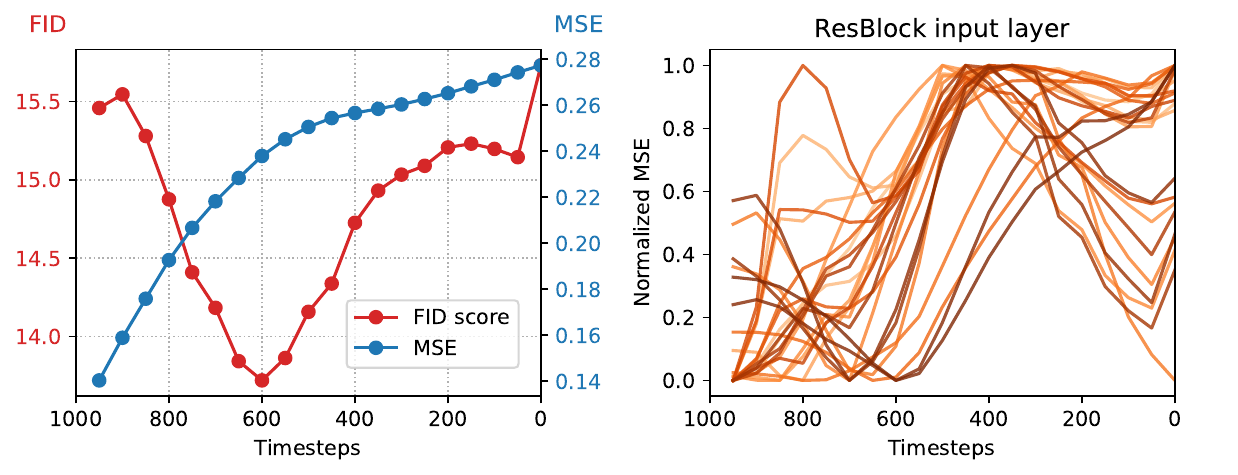}
\end{center}
\vspace{-6mm}
\caption{\textbf{(Left):} Plot of MSE and FID score across timesteps. Quantization noise increases as $t$ decreases and the same amount of noise has a non-uniform impact on the final output. The MSE indicates qauntization error of activations averaged for all layers. 
\textbf{(Right):} Normalized quantization error for different layers over timesteps. Error trends vary across layers along timesteps. }
\vspace{-5mm}
\label{fig:fid_mse}
\end{figure}
}%

\newcommand{\figLayerStat}{%
\begin{figure}[t]
\begin{center}
\includegraphics[width=0.88\columnwidth]{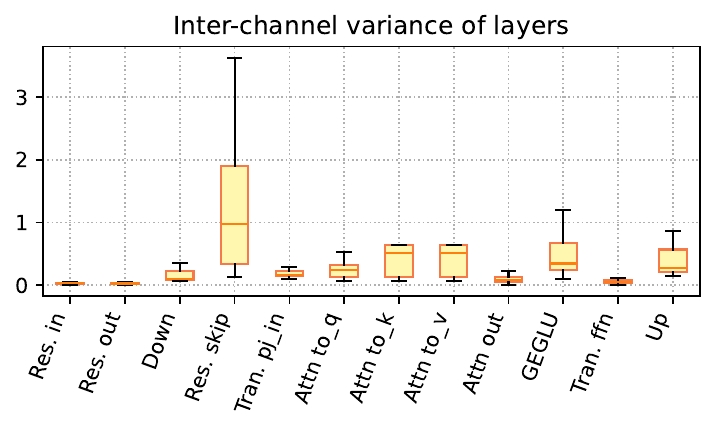}
\end{center}
\vspace{-7mm}
\caption{Inter-channel variance of the activations. \emph{Res.} and \emph{Tran.} indicates the residual block and the transformer layer, respectively. 
}
\vspace{-1.5mm}
\label{fig:layer_stat}
\end{figure}
}%

\newcommand{\figPot}{%
\begin{figure}[t]
\begin{center}
\includegraphics[width=\columnwidth]{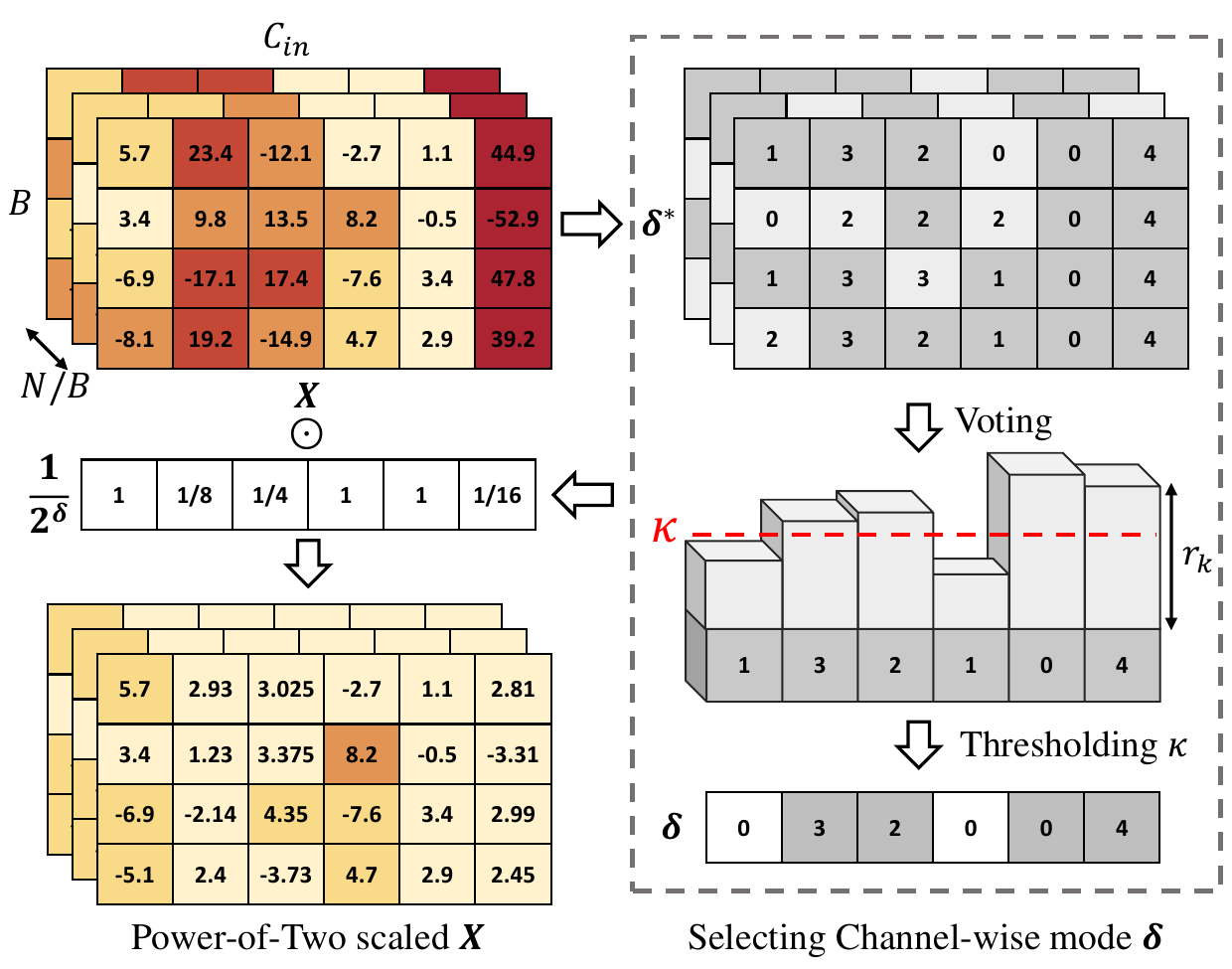}
\end{center}
\vspace{-4mm}
\caption{Illustration of voting algorithm for PTS factor $\delta$. 
}
\vspace{-3mm}
\label{fig:pot}
\end{figure}
}%

\newcommand{\figImagenet}{%
\begin{figure*}[ht]
\begin{center}
\includegraphics[width=\linewidth]{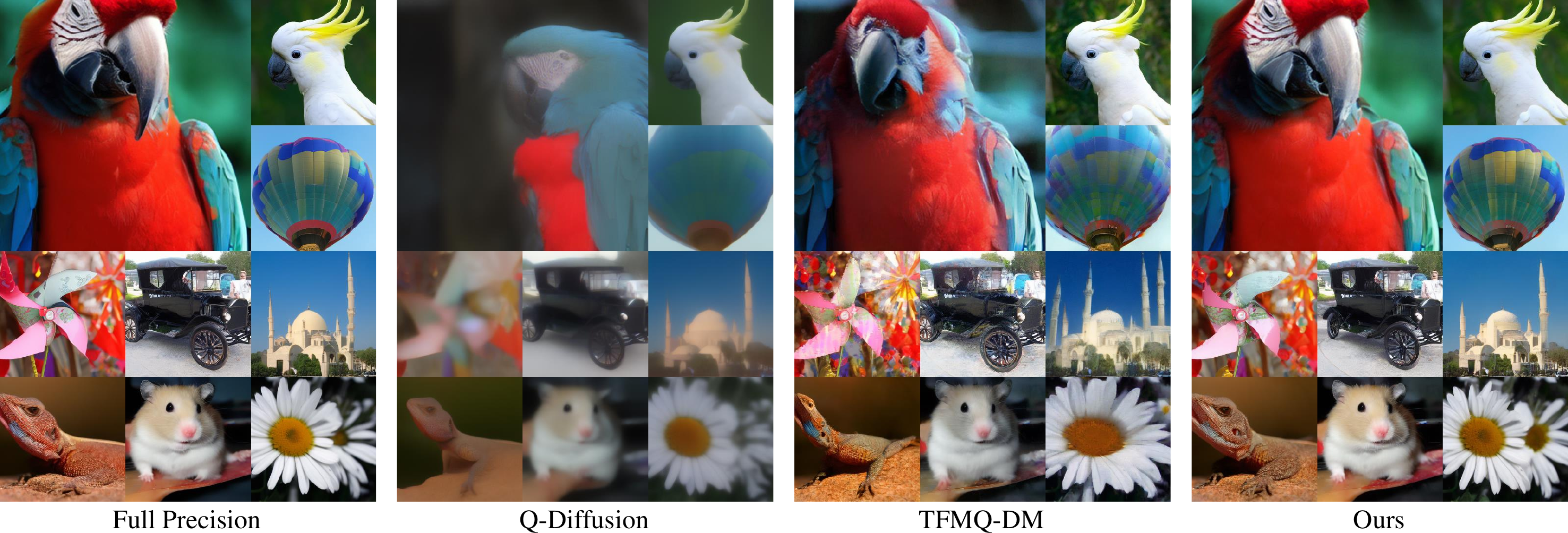}
\end{center}
\vspace{-5mm}
\caption{Visualization of samples on ImageNet 256$\times$256 generated by full precision LDM~\cite{rombach2022high} and W4A6 quantized models using Q-Diffusion~\cite{li2023q}, TFMQ-DM~\cite{huang2024tfmq}, and ours. 
Unlike other methods that often struggle to maintain plausible outputs, the proposed method successfully preserves the high sample quality of full precision models, even under extremely low-bit quantization.
}
\vspace{-4mm}
\label{fig:visualize}
\end{figure*}
}%

\newcommand{\figVsSmoothquant}{%
\begin{figure}[!t]
    \begin{subfigure}[t]{0.47\columnwidth}
        \centering
        \includegraphics[width=\textwidth]{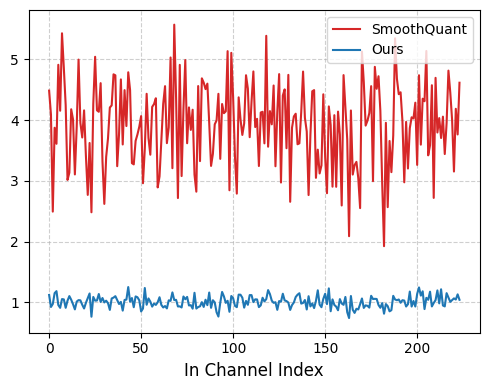}
        \subcaption{Equivalent scaling factor $\tau$}
        \label{fig:vs_smoothquant:a}
    \end{subfigure}
    \begin{subfigure}[t]{0.47\columnwidth}
        \centering
        \includegraphics[width=\textwidth]{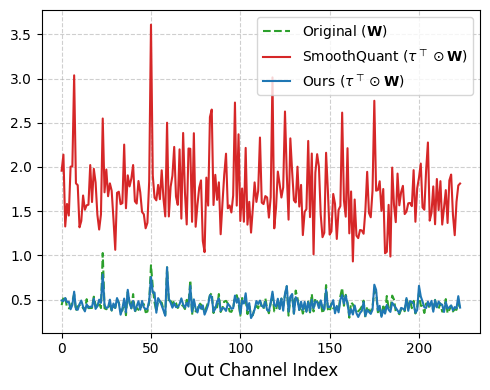}
        \subcaption{Quantization range of $\mathbf{W}$}
        \label{fig:vs_smoothquant:b}
    \end{subfigure}
    \vspace{-3mm}
    \caption{(a) Smoothquant results large equivalent scaling factor. (b) Large scaling factor expands quantization range (Max - Min).}
     \label{fig:vs_smoothquant}
    \vspace{-2mm} 
\end{figure}
}%

\newcommand{\figVisBedroom}{%
\begin{figure*}[!ht]
\begin{center}
\includegraphics[width=\linewidth]{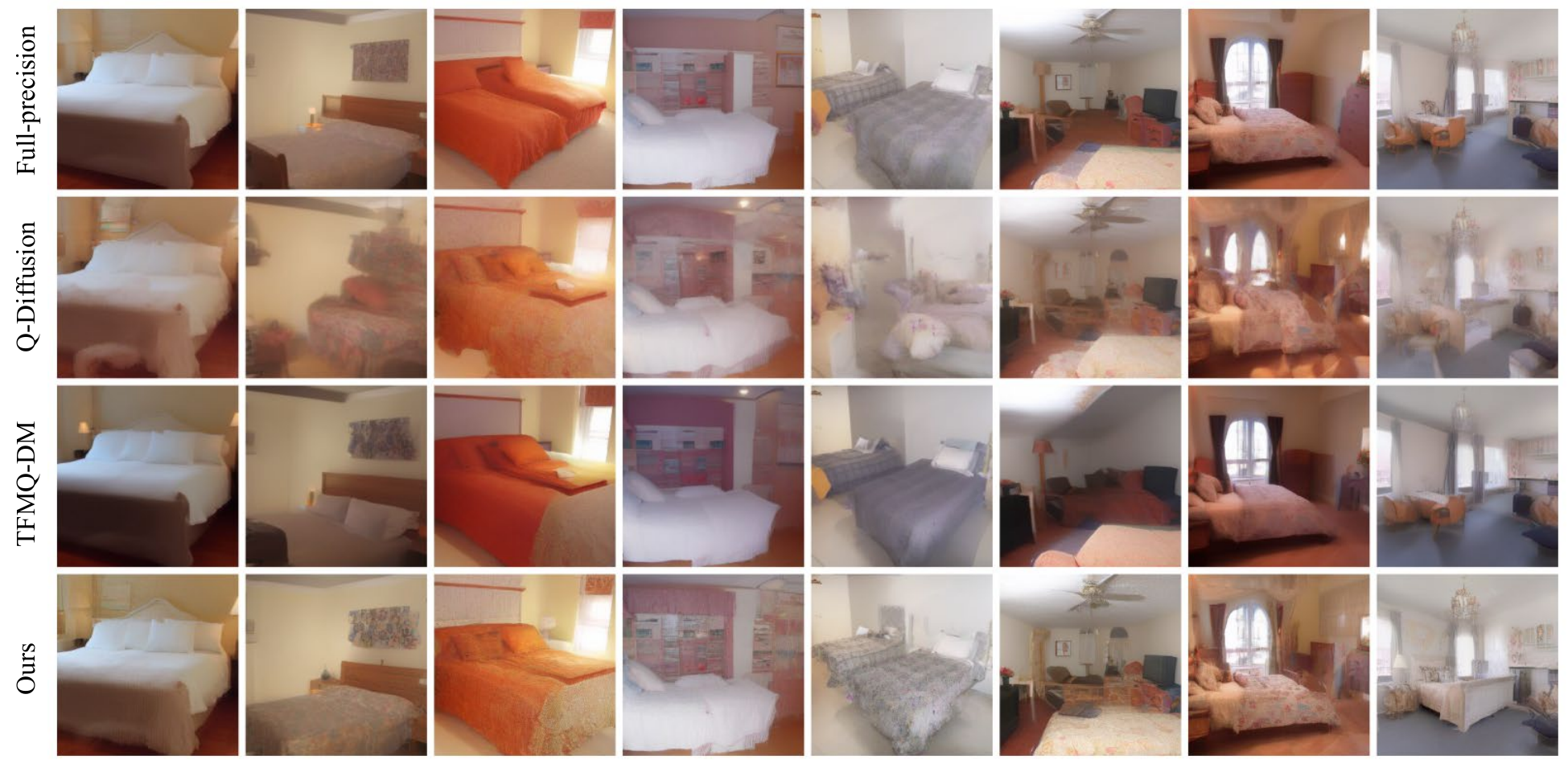}
\end{center}
\caption{Visualization of samples on LSUN-Bedrooms 256$\times$256 generated by full precision LDM~\cite{rombach2022high} and W4A6 quantized models using Q-Diffusion~\cite{li2023q}, TFMQ-DM~\cite{huang2024tfmq}, and ours.}
\label{fig:vis_bedroom}
\end{figure*}
}

\newcommand{\figVisChurch}{%
\begin{figure*}[!ht]
\begin{center}
\includegraphics[width=\linewidth]{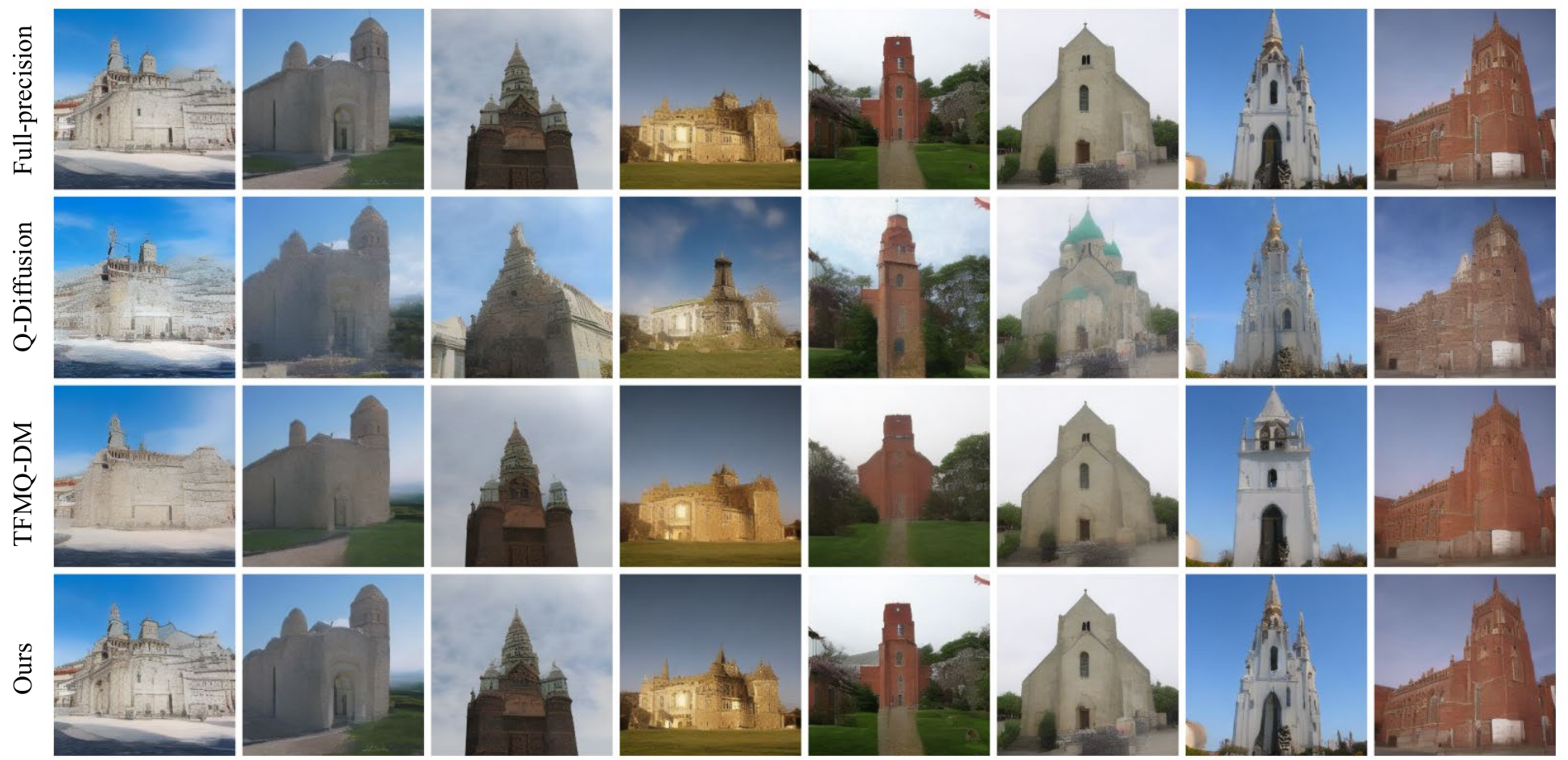}
\end{center}
\caption{Visualization of samples on LSUN-Chruches 256$\times$256 generated by full precision LDM~\cite{rombach2022high} and W4A6 quantized models using Q-Diffusion~\cite{li2023q}, TFMQ-DM~\cite{huang2024tfmq}, and ours.}
\label{fig:vis_church}
\end{figure*}
}

\newcommand{\figVisFFHQ}{%
\begin{figure*}[!ht]
\begin{center}
\includegraphics[width=\linewidth]{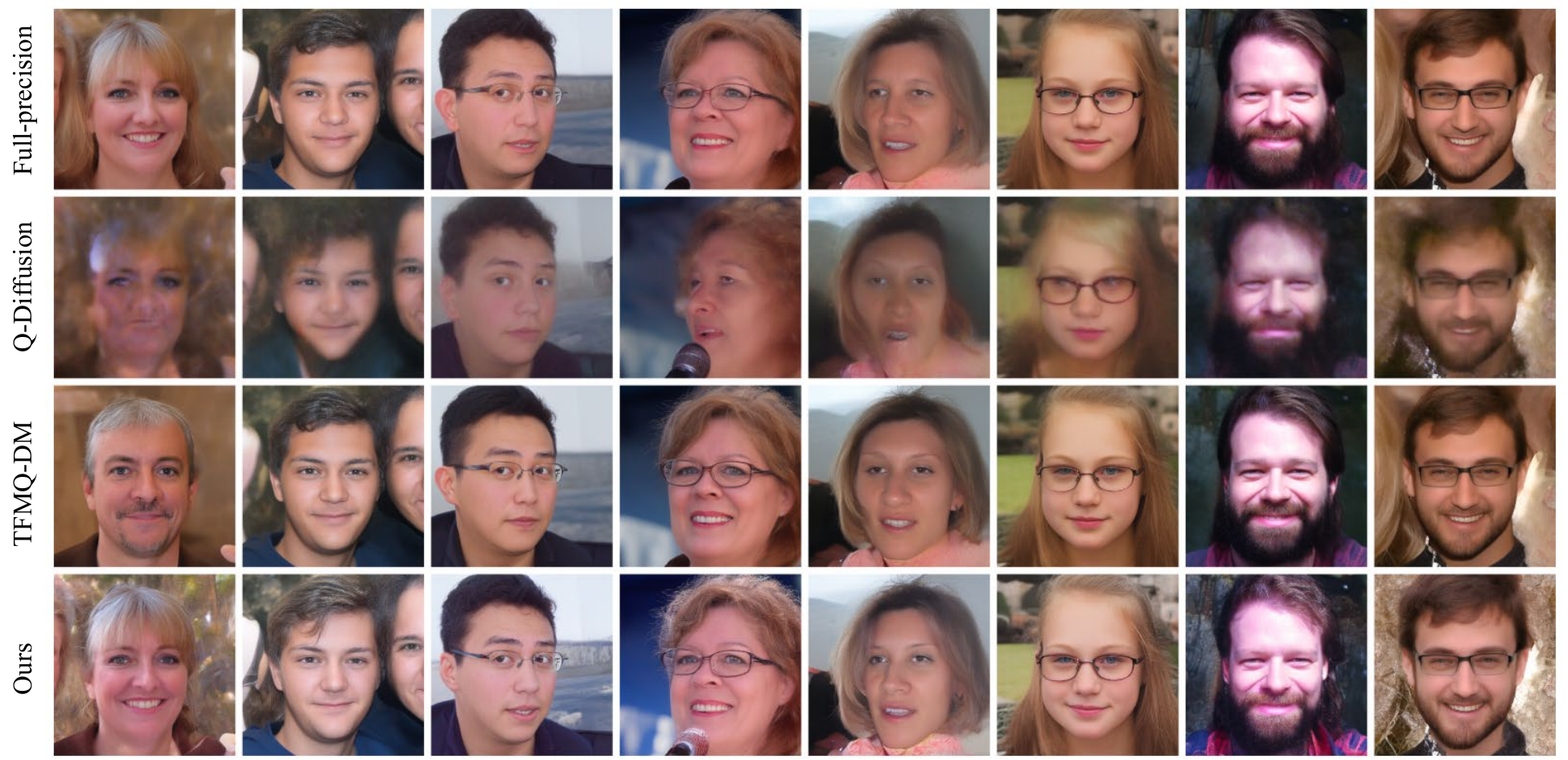}
\end{center}
\caption{Visualization of samples on FFHQ 256$\times$256 generated by full precision LDM~\cite{rombach2022high} and W4A6 quantized models using Q-Diffusion~\cite{li2023q}, TFMQ-DM~\cite{huang2024tfmq}, and ours.}
\label{fig:vis_ffhq}
\end{figure*}
}

\newcommand{\figVisImagenet}{%
\begin{figure*}[!ht]
\begin{center}
\includegraphics[width=\linewidth]{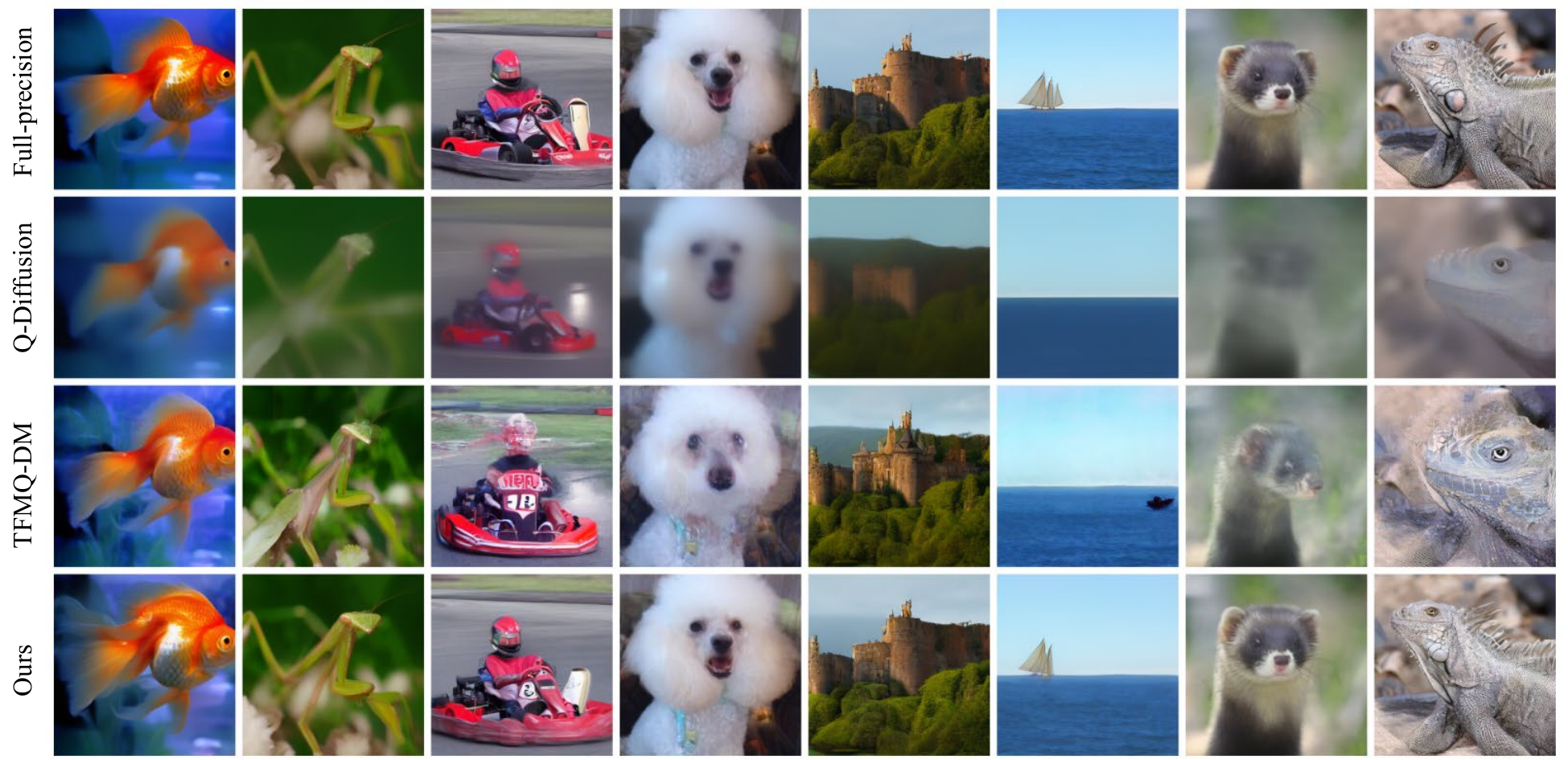}
\end{center}
\caption{Visualization of samples on ImageNet 256$\times$256 generated by full precision LDM~\cite{rombach2022high} and W4A6 quantized models using Q-Diffusion~\cite{li2023q}, TFMQ-DM~\cite{huang2024tfmq}, and ours.}
\label{fig:vis_imagenet}
\end{figure*}
}

\newcommand{\figGemmSpeed}{%
\begin{figure}[!t]
\begin{center}
\includegraphics[width=\linewidth]{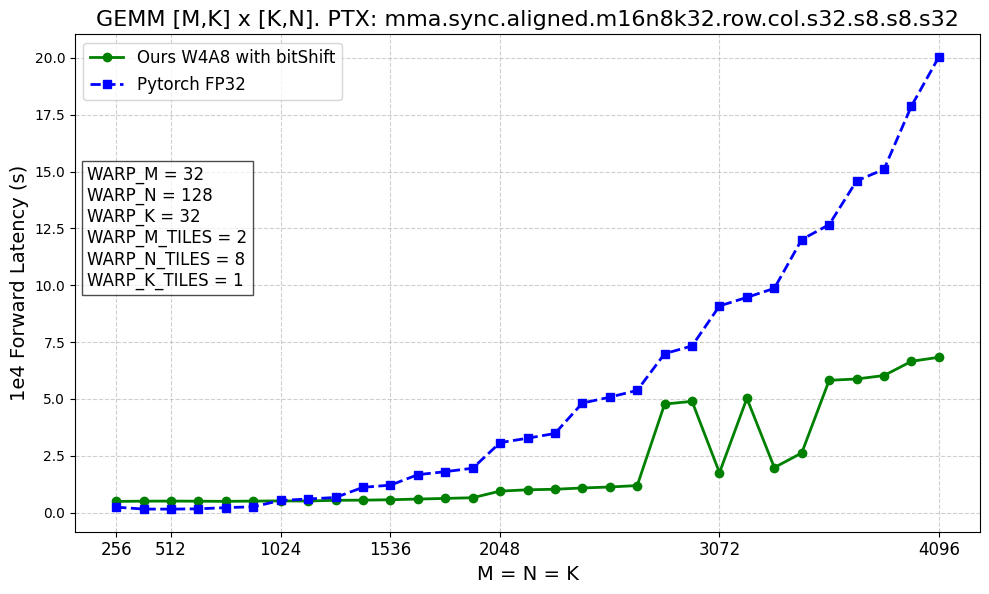}
\end{center}
\caption{Comparison on latency between pytorch fp32 GEMM and our custom W4A8 GEMM kernel including quantization, bit-shifting on weight, GEMM, and dequantization.}
\label{fig:gemm_speed}
\end{figure}
}

\def\paperID{8054} 
\def\confName{ICCV}
\def\confYear{2025}

\title{DMQ: Dissecting Outliers of Diffusion Models for Post-Training Quantization}

\author{Dongyeun Lee \quad Jiwan Hur \quad Hyounguk Shon \quad Jae Young Lee \quad Junmo Kim \\
KAIST \\
\tt\small \{ledoye, jiwan.hur, hyounguk.shon, mcneato, junmo.kim\}@kaist.ac.kr
}

\begin{document}

\twocolumn[{%
\vspace{-2em}
\maketitle%
\figFirstPage%
}]

\begin{abstract}
Diffusion models have achieved remarkable success in image generation but come with significant computational costs, posing challenges for deployment in resource-constrained environments. Recent post-training quantization (PTQ) methods have attempted to mitigate this issue by focusing on the iterative nature of diffusion models. However, these approaches often overlook outliers, leading to degraded performance at low bit-widths. In this paper, we propose a DMQ which combines Learned Equivalent Scaling (LES) and channel-wise Power-of-Two Scaling (PTS) to effectively address these challenges. Learned Equivalent Scaling optimizes channel-wise scaling factors to redistribute quantization difficulty between weights and activations, reducing overall quantization error. Recognizing that early denoising steps, despite having small quantization errors, crucially impact the final output due to error accumulation, we incorporate an adaptive timestep weighting scheme to prioritize these critical steps during learning. 
Furthermore, identifying that layers such as skip connections exhibit high inter-channel variance, we introduce channel-wise Power-of-Two Scaling for activations. 
To ensure robust selection of PTS factors even with small calibration set, we introduce a voting algorithm that enhances reliability. Extensive experiments demonstrate that our method significantly outperforms existing works, especially at low bit-widths such as W4A6 (4-bit weight, 6-bit activation) and W4A8, maintaining high image generation quality and model stability. 
The code is available at \href{https://github.com/LeeDongYeun/dmq}{https://github.com/LeeDongYeun/dmq}.
\end{abstract}    
\vspace{-8mm}
\section{Introduction}
\label{sec:intro}
\vspace{-2mm}
Diffusion models \cite{ho2020denoising,nichol2021improved,rombach2022high,peebles2023scalable} have recently shown remarkable success in synthesizing high-fidelity images, enabling various applications such as image editing \cite{lugmayr2022repaint,li2022srdiff,zhang2023adding,ruiz2023dreambooth,hur2024expanding,shi2024dragdiffusion}, 3D generation \cite{poole2023dreamfusion,ruiz2023dreambooth,lin2023magic3d,yu2023text}, and video \cite{ho2022video,jabri2022scalable,singer2022make,blattmann2023align} generation.
However, despite their impressive generative capabilities, diffusion models face significant computational challenges due to the iterative nature of the denoising process, which can require hundreds or even thousands of steps.
Consequently, accelerating this process while preserving output quality is essential for practical applications and has become a key research focus in the field.

Quantization \cite{Jacob_2018_CVPR,nagel2021white,gholami2022survey} reduces computational and memory demands by lowering the bit-precision of neural networks.
However, quantization of diffusion models presents unique challenges due to their iterative nature. 
The iterative nature results in highly varying activation distributions across multiple timesteps and quantization error accumulation as denoising progresses, hindering accurate quantization.
To overcome this, several methods have explored Post-Training Quantization (PTQ) for diffusion models, focusing on composition of calibration data considering multiple timesteps \cite{shang2023post,li2023q,wang2024towards,liu2024enhanced}, adapting quantization parameters across timesteps \cite{so2024temporal,huang2024tfmq} or correcting noise within the diffusion process \cite{he2024ptqd,yao2024timestep}.
However, they often overlook channel-wise outliers which stretch the quantization range, hindering effective quantization of non-outlier channels.
As a result, existing methods often exhibit large performance degradation in low-bit settings (e.g. W4A6).

In this work, we address the challenge of outliers in diffusion model quantization.
A natural approach to mitigating outliers is Equivalent Scaling—originally introduced for LLMs quantization \cite{wei2023outlier,lin2024awq,xiao2023smoothquant}—which reduces quantization difficulty by transferring the challenges of outlier channels in activations to non-outlier channels in weights and vice versa. 
This involves simultaneously dividing each activation channel and multiplying the corresponding weights by a channel-specific scaling factor. 
Although this approach has shown promise in LLMs, its naive application often fails for several reasons.
First, inaccurate scaling factors can drastically degrade output quality.
As shown in \cref{tab:vs_smoothquant}, SmoothQuant \cite{xiao2023smoothquant}, which determines scaling factor based on the ratio of maximum activation and weight magnitudes, results in poor performance for diffusion models.
Since diffusion models typically exhibit much larger activations than weights, SmoothQuant yields large scaling factor (\cref{fig:vs_smoothquant:a}), which significantly increases quantization range of weight and in turn amplifies weight quantization errors(\cref{fig:vs_smoothquant:b}). 
In diffusion models, weights are used at every sampling step, making inaccurate weight quantization severely degrade performance, with errors compounding over iterations.
Second, Equivalent Scaling redistributes outliers between activations and weights but does not eliminate outliers. 
By scaling activations down and compensating with an inverse scaling of weights, this method reduces the overall quantization difficulty.
However, it inherently shifts the burden rather than removing it.
Consequently, layers exhibiting extremely large outliers remain problematic.

\vspace{-1.5mm}
To address these issues, we propose DMQ, a new post-training quantization (PTQ) framework tailored for diffusion models.
DMQ unifies two key techniques: \textbf{L}earned \textbf{E}quivalent \textbf{S}caling (\textbf{LES}) and channel-wise \textbf{P}ower-of-\textbf{T}wo \textbf{S}caling (\textbf{PTS}), which together ensure accurate quantization even under low-bit constraints.
First, LES learns accurate scaling factors that balance quantization difficulty between activation and weight by minimizing overall quantization error.
We observe that early denoising steps, despite having smaller quantization errors, significantly influence the final output quality. 
To account for this, we introduce an adaptive timestep weighting loss that prioritizes learning in critical timesteps, enhancing overall quantization performance.
Second, to handle extreme outliers that remain unresolved by Equivalent Scaling, we introduce PTS.
Unlike LES, which compensates for reduced activation difficulty by increasing weight difficulty and vice versa, PTS directly removes extreme activation outliers by scaling them with power-of-two factors.
These scaling factors are efficiently processed during weight loading via bit-shifting operations, ensuring that the quantization difficulty is not merely transferred but instead removed with minimal overhead.
To prevent overfitting on small calibration datasets, we introduce an effective voting algorithm which selects optimal scaling factors based on statistical consensus across calibration samples, enhancing robustness and reliability.

LES, which is applied across all layers without introducing extra overhead, finely adjusts outliers using floating-point scaling factors. 
By contrast, PTS is selectively used only for layers with extremely large outliers (e.g. skip connection), incurring a small overhead in the form of bit-shifting but enabling drastic outlier removal through power-of-two scaling.
By combining the strengths of LES and PTS, our approach effectively addresses the challenges of outliers in diffusion model quantization. 
Extensive experiments demonstrate that DMQ consistently achieves superior performance across various datasets and architectures. 
Notably, we achieve stable 4-bit weight and 6-bit activation (W4A6) quantization—where previous methods often fail—while preserving high image generation quality.

\figVsSmoothquant
\tabVsSmoothquant
\vspace{-2mm}
\section{Related work}
\label{sec:rw}
\vspace{-2mm}
\noindent\textbf{Quantization.} Quantization is a core technique for reducing memory overhead and accelerating inference by converting high-precision floating-point values into lower-bit integer approximations.
Of the two primary quantization methods — Quantization-Aware Training (QAT) \cite{esser2019learned,choi2018pact,jung2019learning} and Post-Training Quantization (PTQ) \cite{Nagel_2019_ICCV,nagel2020up,li2021brecq,lee2018quantization} — PTQ stands out for its efficiency in terms of time and data, as it bypasses the need for retraining.
However, traditional methods often fail to accurately quantize diffusion models, as they require multiple denoising steps with a single model.

\noindent\textbf{Quantization of diffusion models.} Recently, various methods have explored quantization for diffusion models, enabling low-bit quantization while preserving performance \cite{he2024efficientdm,NEURIPS2023_f1ee1cca,shang2023post,li2023q,he2024ptqd,so2024temporal,huang2024tfmq,yao2024timestep,wang2024towards,liu2024enhanced}.
In QAT approach, Q-DM \cite{NEURIPS2023_f1ee1cca} proposes timestep-aware smoothing operation for attention. EfficientDM \cite{he2024efficientdm} introduces a quantization-aware low-rank adapter (LoRA) that is jointly quantized into low-bit. 
However, QAT requires intensive computation and data resources, making it less practical than PTQ.

Research in PTQ for diffusion models has largely branched into three directions: calibration data composition \cite{shang2023post,li2023q,wang2024towards}, noise correction \cite{he2024ptqd,yao2024timestep}, and timestep-specific quantization parameters \cite{huang2024tfmq,so2024temporal}.
To enhance calibration data, PTQ4DM \cite{shang2023post} samples calibration data from a normal distribution across timesteps, Q-Diffusion \cite{li2023q} employs uniform sampling, and EDA-DM \cite{liu2024enhanced} enhances alignment of distribution with original samples.
Addressing quantization noise, PTQD \cite{he2024ptqd} proposes to correct quantization noise correlated with the denoising process, and TAC-Diffusion \cite{yao2024timestep} presents a timestep-aware correction to mitigate errors throughout the denoising process. 
Alternatively, TFMQ-DM \cite{huang2024tfmq} applies different quantization parameters across timesteps, whereas TDQ \cite{so2024temporal} proposes to learn these parameters.
Despite the advancements of these methods, they often suffer from significant performance degradation under 8-bit activation because they overlook the outliers.

\figBeforeAfter

\noindent\textbf{Equivalent channel-wise scaling.}  In a neural network, activations often have varying distributions across channels, making it challenging to accurately represent them by per-tensor quantization.
This issue is particularly pronounced in large language models (LLMs), where extreme outliers in specific channels lead to a wide activation range, hindering accurate quantization \cite{yao2022zeroquant,dettmers2022gpt3,wei2022outlier}.
To address this, several methods adopt equivalent channel-wise scaling, which scales the activations channel-wise and compensates for this scaling by inversely scaling the weights \cite{wei2023outlier,xiao2023smoothquant,lin2024awq}. This approach maintains mathematical equivalence while compacting the activation range within each channel, facilitating finer quantization without loss of precision due to outliers.

Building on the success in LLMs, several recent studies have adopted Equivalent Scaling for DiT \cite{peebles2023scalable} quantization \cite{zhao2024vidit,wu2024ptq4dit}.
Although they have achieved some success in DiT—where adaptation is straightforward due to their similarity to LLMs—they are based on SmoothQuant \cite{xiao2023smoothquant} which results suboptimal performance in Stable Diffusion (\cref{tab:vs_smoothquant}).
Additionally, ViDiT-Q \cite{zhao2024vidit} has practical limitations, as the scaling factors vary with each timestep and therefore cannot be fused into the weights.
Unlike these approaches, our method learns these factors to efficiently capture the dynamically changing activations across timesteps.
\vspace{-6mm}
\section{Preliminaries}
\label{sec:prelimiaries}
\vspace{-2mm}
\noindent\textbf{Diffusion models}
\cite{sohl2015deep,ho2020denoising} are a family of generative models that approximate the data distribution $p(\vx_0)$ by denoising process.
Specifically, the forward Markov chain $q(\vx_t|\vx_{t-1})$ perturbs the data with Gaussian noise and is defined as
\vspace{-1mm}
\begin{equation}
    q(\vx_t|\vx_{t-1}) = \mathcal{N}\left(\vx_t;\sqrt{1-\beta_t}\vx_{t-1}, \beta_t\mathbf{I}\right),
\vspace{-1mm}
\end{equation}
where $\beta_t$ is a noise schedule that is a strictly decreasing function of time $t\in[0,T]$. Then, the learned reverse process aims to gradually recover the clean data $\vx_0$ from the noised data $\vx_t$ to generate novel samples:
\vspace{-1mm}
\begin{equation}
    p_\theta(\vx_{t-1}|\vx_t) = \mathcal{N}\left(\vx_{t-1};\mu_\theta(\vx_t, t), \Sigma_\theta(\vx_t,t)\right),
\vspace{-1mm}
\end{equation}
where the $\mu_\theta$ and $\Sigma_\theta$ are neural networks that predict mean and variance of input $\vx_t$ given timestep $t$~\cite{nichol2021improved}.

To efficiently implement the iterative denoising process, diffusion models share the parameters across timesteps $t$ using the position embedding~\cite{vaswani2017attention}. 
This often raises challenges for quantizing diffusion models, as activation varies across timestep conditioning \cite{shang2023post,li2023q}.
Furthermore, quantization errors are introduced at each denoising step, which can compound over time \cite{li2023q,huang2024tfmq}. 
This accumulation of errors throughout the sampling process leads to the model drifting away from the training distribution, resulting in significant degradation of the final output quality \cite{li2023alleviating,ning2023elucidating,daras2024consistent}.

\noindent\textbf{Quantization.}
Among some quantization methods, we focus on uniform quantization in this work.
Given a tensor $\mathbf X$, its quantization and dequantization functions are defined as:
\begin{equation}
\vspace{-1mm}
    \mathbf{\tilde{X}} = \text{clamp}\left(\left\lfloor \frac{\mathbf{X}}{s} \right\rceil , l, u\right), \quad \mathbf{X} \approx Q(\mathbf{X}) = s \cdot \mathbf{\tilde{X}},
\label{eq:base_q}
\vspace{-1mm}
\end{equation}
where $s$ is the scale factor, and $\lfloor\cdot\rceil$ denotes round-to-nearest-integer operator.
$\text{clamp} (\cdot)$ truncates values to $[l, u]$, yielding a low-bit representation $\mathbf{\tilde{X}}$ of $\mathbf{X}$. Here, the range $[l, u]$ is determined by the bit-width of the quantization. When $s$ is a vector, $\mathbf{X} / s$ implies channel-wise division.

Consider a linear layer $\mathbf{Y} = \mathbf{X}\mathbf{W} \approx (s^{(\mathbf{X})} \mathbf{\tilde{X}})(s^{(\mathbf{W})} \mathbf{\tilde{W}})$, where $\mathbf{X} \in \mathbb{R}^{B\times C_{in}}$ is the activation, $\mathbf{W} \in \mathbb{R}^{C_{in}\times C_{out}}$ is the weight, and $\mathbf{Y} \in \mathbb{R}^{B\times C_{out}}$ is the output. $s^{(\mathbf{X})}$ and $s^{(\mathbf{W})}$ denote the scales for $\mathbf{X}$ and $\mathbf{W}$, respectively.
Note that the following discussion applies equally to convolutional layers, but for simplicity, we will illustrate the concepts with a linear layer.
To enable efficient integer matrix multiplication, it is necessary to factor out the scale $s$, which is a high-precision value, from the summation. 
Under this condition, the matrix multiplication can be formulated as:
\vspace{-1mm}
\begin{equation}
    \mathbf{Y}_{ij} \approx s_{i}^{(\mathbf{X})} s_{j}^{(\mathbf{W})} \left(\sum_{k=1}^{C_{in}} \mathbf{\tilde{X}}_{ik} \cdot \mathbf{\tilde{W}}_{kj} \right) .
\vspace{-1mm}
\end{equation}
This formulation shows that integer matrix multiplication is feasible when activations are quantized with per-tensor or per-sample (row) granularity and weights are quantized in per-tensor or per-channel (column) granularity.
Since per-sample quantization requires online computation, per-tensor quantization is generally used for activations.

\vspace{-3mm}
\section{Method}
\label{sec:method}
\vspace{-1mm}
\subsection{Learned Equivalent Scaling}
\vspace{-2mm}
\noindent\textbf{Channel-wise equivalent scaling.}
Our goal is to reduce quantization errors by optimizing channel-wise scaling factors $\tau \in \mathbb{R}^{C_{in}}$.
We reformulate the matrix multiplication to incorporate these scaling factors:
\begin{equation}
    \mathbf{Y} = \left(\mathbf{X} / \tau\right) \left(\tau^\top \odot \mathbf{W}\right) = \mathbf{\hat{X}} \mathbf{\hat{W}},
\label{eq:equivalent}
\end{equation}
where $/$ and $\odot$ denote channel-wise division and multiplication, respectively.
By introducing $\tau$, outliers are bidirectionally redistributed, resulting in scaled versions $\mathbf{\hat{X}}$ and $\mathbf{\hat{W}}$ that are more amenable to quantization.

The optimization objective to find the optimal $\tau$ for the $i$-th sample is defined as:
\vspace{-1mm}
\begin{equation}
    \mathcal{L}_i = \bigl\Vert\mathbf{X}_i \mathbf{W} - Q(\mathbf{\hat{X}}_i)Q(\mathbf{\hat{W}})\bigr\Vert^2,
\vspace{-1mm}
\end{equation}
where $Q(\cdot)$ denotes standard MinMax quantizer \cite{nagel2021white}.
Minimizing $\mathcal{L}_i$ ensures that the quantized outputs closely approximate the original outputs.
As shown in the \cref{fig:before_after}, the quantization difficulties caused by outliers in weights and activations are bidirectionally redistributed to non-outlier elements, thereby reducing the overall quantization error.

\figFidMse

\noindent\textbf{Analysis on quantization error.} 
In a diffusion model, the activation distribution varies dynamically across timesteps, and quantization errors are accumulated over the denoising process \cite{li2023q,huang2024tfmq}. 
Therefore, before directly optimizing the objective loss across all timesteps, it is crucial to carefully analyze the behavior of quantization error over timesteps.
To investigate this, we initially quantize the network using a MinMax quantizer and measure average quantization error across all layers.
As illustrated in \cref{fig:fid_mse}, the quantization error increases as denoising progresses (i.e., as the timestep $t$ decreases).
This observation raises a central question: \emph{does a higher quantization error at a specific timestep correlate with a greater degradation of the final image quality?}

To answer this, we probed the diffusion model by injecting small random noise $\mathcal{U}(0, 0.1)$ at a specific timestep to activations of the quantized model and measuring the FID  \cite{heusel2017gans} of the generated images. 
As shown in \cref{fig:fid_mse}, the final output quality does not directly correlate with the quantization error across timesteps. 
Specifically, while the later timesteps (smaller $t$) exhibit larger quantization errors and significantly degraded FID scores, the earlier timesteps (large $t$) also substantially affect the final image quality despite having smaller quantization errors. 
This is due to the fact that errors at early timesteps accumulate throughout the denoising process, ultimately impacting the final image quality.
These findings highlight the necessity of carefully quantizing the model at early timesteps, even if the immediate quantization error appears small.

\noindent\textbf{Adaptive timestep weighting.}
Applying a uniform loss across timesteps is suboptimal, as later steps typically exhibit larger errors, biasing optimization. 
Moreover, as shown in the \cref{tab:ablation_weighting}, simple monotonic weighting that prioritizes early steps performs even worse.
While the average error increases as $t$ decreases (\cref{fig:fid_mse} left), individual layers exhibit varying trends, with some having minimal error even at later timesteps (\cref{fig:fid_mse} right).
Thus, naively prioritizing early steps can overlook critical later steps in certain layers.
To address this, we propose an adaptive timestep weighting that dynamically adjusts weights based on timesteps and the observed quantization error.
Our adaptive loss function for each batch $B$ is formulated as:
\vspace{-1mm}
\begin{equation} 
    \mathcal{L} = \frac{1}{B} \sum_{i=1}^{B} \lambda_{t_i}\mathcal{L}_i, 
\vspace{-1mm}
\end{equation}
where $\lambda_{t_i}$ adjusts the contribution of each sample based on its timestep $t_i$. 
The weighting factor $\lambda_{t_i}$ is defined as:
\vspace{-1mm}
\begin{equation} 
\lambda_{t_i} = \left( 1 - \frac{\Lambda_{t_i}}{\sum_{t^{\prime} \in T} \Lambda_{t^{\prime}}} \right)^{\alpha}, 
\vspace{-1mm}
\end{equation}
where $T$ is set of timesteps used for calibration and $\Lambda_{t}$ is accumulated loss for the timestep $t$.
This design is inspired by focal loss \cite{lin2017focal}, which emphasizes harder-to-classify samples by assigning larger weights. Similarly, our approach prioritizes critical timesteps with lower accumulated errors.
Specifically, a higher $\alpha$ shifts optimization toward early denoising steps (\cref{fig:fid_mse} left), where small quantization errors can still significantly impact final output quality due to error accumulation.
This adaptive weighting scheme balances both quantization error magnitude and its influence on the denoising trajectory.
The accumulated loss $\Lambda_{t}$ is updated iteratively as follows:
\vspace{-1mm}
\begin{equation} 
\Lambda_t \leftarrow \xi \Lambda_t + (1 - \xi) \mathbb{E}_{i \in I_t}[\mathcal{L}_i],
\vspace{-1mm}
\end{equation}
where $I_t = \{i| t_i = t\}$ is the set of samples at timestep $t$.
Using a moving average with momentum $\xi = 0.95$, we prevent abrupt changes in weights, ensuring stability while capturing shifts in quantization error throughout training.

\figLayerStat

\noindent\textbf{Fusing $\tau$ with static parameters for efficient inference.}
Applying LES factors $\tau$ in \cref{eq:equivalent} directly during inference introduces additional overhead due to the need for online computation. 
Whereas $\tau$ can be fused into the weight $\mathbf{W}$ since they are static, fusing $\tau$ directly with activation $\mathbf{X}$ is not feasible as they change dynamically. 
Previous works on LLMs \cite{wei2023outlier,xiao2023smoothquant,lin2024awq} address this by fusing $\tau$ into previous linear operations (e.g., layer norm, linear layer).
However, most diffusion models \cite{ho2020denoising,rombach2022high} have a nonlinear operation \cite{ramachandran2017swish} before matrix multiplication, preventing such fusion.
To overcome this, we incorporate $\tau$ into the quantization process by using a modified scale for activation: 
\vspace{-2mm}
\small{
\begin{align}
    \mathbf{\tilde{X}} &= \text{clamp}\left(\left\lfloor \frac{\mathbf{X}} { \{\tau \odot s^{(\mathbf{X})} \} } \right\rceil , l, u \right), \mathbf{\hat{X}} \approx   s^{(\mathbf{X})} \cdot \mathbf{\tilde{X}} \\
    \mathbf{\tilde{W}} &= \text{clamp}\left(\left\lfloor \frac{ \{ \tau^\top \odot \mathbf{W} \} }{s^{(\mathbf{W})}} \right\rceil , l, u \right), \mathbf{\hat{W}} \approx s^{(\mathbf{W})} \cdot \mathbf{\tilde{W}},
\vspace{-1mm}
\end{align}
}\normalsize
where $\{\cdot\}$ indicates pre-computed values.
For weights, $\tau$ is fused directly into the weight values; in contrast, for activations, $\tau$ is integrated into the scale $s^{(\mathbf{X})}$ during quantization. During dequantization, both weights and activations use original scales $s^{(\mathbf{X})}$ and $s^{(\mathbf{W})}$, respectively. This is feasible as our approach employs static quantization, ensuring that  $s^{(\mathbf{X})}$ remains fixed.
This allows us to apply the scaling factors without additional overhead during inference.

\figPot

\setlength{\textfloatsep}{5pt}
\begin{algorithm}[!t]
\caption{Voting Algorithm for PTS Factor}
\label{alg:channelwise_scaling}
\begin{algorithmic}
\State \textbf{Input:} calibration data $\{\mathbf{X}_i\}^N_{i=1}$, number of candidate scales \( D \), agreement threshold \( \kappa \), per-tensor scale \(s\)

\Comment{Candidate Selection by Per-Channel Quantization}
\For{sample index $i \in \{1,...,N\}$}
    \ForAll {channel index \(k \in \{1, \dots , C_{in}\}\)}
        \State \(\delta_{i,k}^* \leftarrow \argmin_{d \in \{0, \dots , D\}} \bigl\Vert\mathbf{X}_{ik} - Q_{(s \cdot 2^d)}(\mathbf{X}_{ik})\bigr\Vert^2\)
    \EndFor
\EndFor
\\

\Comment{Robust Voting Mechanism by thresholding}
\ForAll{channel index \(k \in \{1, \dots , C_{in}\}\)}
    \State \( \delta_k^{\text{mode}} = \text{mode} \left( \{ \delta_{i,k}^* \}_{i=1}^{N} \right) \) 
    \State \( r_k = \frac{1}{N} \sum_{i=1}^N \mathbf{1}\left\{\delta_{i,k}^* = \delta_k^{\text{mode}}\right\} \)
    \If{ \( r_k > \kappa \) }
        \State \( \delta_k \leftarrow \delta_k^{\text{mode}} \)
    \Else
        \State \( \delta_k \leftarrow 0 \) 
    \EndIf
\EndFor

\State \Return PTS factors \( \delta = [\delta_1, \dots, \delta_{C_{in}}] \)
\end{algorithmic}
\end{algorithm}

\tabUncond

\subsection{Power-of-Two Scaling}
\noindent\textbf{Challenges with large outliers in activations.}
Upon examining activation distributions in the diffusion model, we found that some layers—especially skip connections lacking normalization—exhibit significant inter-channel variance due to extreme outliers (\cref{fig:layer_stat}). Notably, while previous work \cite{li2023q} focused on bimodal distributions in UNet shortcut layers, our analysis shows that ResBlock skip connections suffer from even more severe outlier-induced variance. This high inter-channel variance poses substantial challenges for quantization. Although LES reduces quantization errors by redistributing difficulty between weights and activations, it alone cannot fully mitigate these extreme cases, which pose substantial challenges for quantization.

\noindent\textbf{Power-of-Two quantization of activations.}
To address these challenges, we introduce a channel-wise Power-of-Two Scaling (PTS) for activations. 
This approach scales activations on a per-channel basis using power-of-two scaling factors, effectively adjusting the dynamic range of each channel individually. The scaling is formulated as:
\vspace{-2mm}
\begin{equation}
    \mathbf{\tilde{X}} = \text{clamp}\left(\left\lfloor \frac{\mathbf{X}}{2^{\boldsymbol{\delta}} \odot s^{(\mathbf{X})}} \right\rceil , l, u \right)
    \vspace{-2mm}
\end{equation}
where $\boldsymbol{\delta} \in \mathbb{R}^{C_{in}}$ contains channel-wise power-of-two exponents, and $2^{\boldsymbol{\delta}} = [2^{\delta_1}, 2^{\delta_2}, \dots, 2^{\delta_{C_{in}}}]$.
Note that PTS applies only to activations, while weights follow \cref{eq:base_q}.
Integration of PTS into matrix multiplication is formulated as:
\vspace{-2mm}
\small{\begin{align}
    \mathbf{Y}_{ij} &\approx s_{i}^{(\mathbf{X})} s_{j}^{(\mathbf{W})} \left(\sum_{k=1}^{C_{in}} 2^{\delta_{k}} \tilde{\mathbf{X}}_{ik} \cdot\tilde{\mathbf{W}}_{kj} \right) \nonumber \\
    &= s_{i}^{(\mathbf{X})} s_{j}^{(\mathbf{W})} \left(\sum_{k=1}^{C_{in}} \tilde{\mathbf{X}}_{ik} \cdot\left(\tilde{\mathbf{W}}_{kj} \ll \boldsymbol{\delta}_{k}\right) \right),
    \label{eq:bitshift}
    \vspace{-2mm}
\end{align}
}\normalsize
where $\ll$ denotes the left-bit-shift operation. 
\cref{eq:bitshift} shows that multiplying by a PTS factor is equivalent to shifting the bits of the product, which is highly efficient on hardware \cite{armeniakos2022hardware,you2020shiftaddnet,elhoushi2021deepshift,you2024shiftaddllm}.
In practice, bit-shifting is applied immediately after loading weights at kernel execution (rather than during multiply-accumulation), ensuring minimal overhead.

Selecting an appropriate PTS factor $\boldsymbol{\delta}$ is crucial for minimizing quantization error. If $\boldsymbol{\delta}$ is too small, large salient activations are not sufficiently scaled down, pushing them outside the quantization range and causing severe clamping errors. In contrast, if $\boldsymbol{\delta}$ is too large, the activations are overly compressed, increasing the rounding errors.
A straightforward approach is to select $\boldsymbol{\delta}$ values that minimize quantization error over the calibration set.
However, we found that this degrades the model’s performance on unseen data, making it unsuitable for PTQ, where only a small calibration set is available.
To overcome this, we propose a robust voting algorithm for selecting the PTS factors, which remains effective even with a small calibration set.

\figImagenet

\noindent\textbf{Robust voting algorithm for scaling factor selection.}
As illustrated in \cref{fig:pot}, the voting algorithm for PTS factors comprises the following two major components.
\begin{itemize}
\item \emph{Candidate selection by per-channel quantization}.
For each calibration sample and channel, we evaluate candidate power-of-two scaling factors $\{2^0, 2^1, ..., 2^D\}$ and select the factor $\delta_{i,k}^{*}$ that minimizes the quantization error.
\item \emph{Robust voting mechanism by thresholding}.
After determining the best scaling factors per sample and channel, we aggregate them across all samples by computing the mode $\delta_{k}^{\text{mode}}$ for each channel $k$. We then compute the agreement ratio $r_k$ , the fraction of samples that select $\delta_{k}^{\text{mode}}$. if $r_k > \kappa$, we assign $\delta_{k} =\delta_{k}^{\text{mode}}$; otherwise, we default to $\delta_{k} = 0$, meaning no scaling is applied.

\end{itemize}
This conservative approach prevents significant activation distortion by rejecting scaling factors with low agreement, thereby avoiding excessive clamping errors that arise when inappropriate PTS factors push activations beyond their typical quantization range. 
By requiring consensus, the method mitigates overfitting to anomalies in small calibration datasets and ensures robust scaling factor selection. 
Algorithm~\ref{alg:channelwise_scaling} summarizes the full procedure.
\vspace{-2mm}
\section{Experiments}
\vspace{-2mm}

\noindent\textbf{Implementation details.}
We apply channel-wise quantization for weights and static tensor-wise quantization for activations. For a fair comparison, all methods quantize each layer except the input and output layers, which remain in full precision per common practice\cite{esser2019learned}.
Equivalent scaling is optimized in a layer-wise manner, inspired by AdaRound \cite{nagel2020up}.
After learning the equivalent scaling, we employ BRECQ \cite{li2021brecq} for weight quantization. 
The setup of the calibration data follows that of Q-Diffusion \cite{li2023q}, except for Stable Diffusion. 
For Stable Diffusion, we do not employ classifier-free guidance when sampling the calibration set following TAC \cite{yao2024timestep}.
Additionally, instead of quantizing time embedding layers, we cache their output values, which is more efficient and effective than direct quantization \cite{sui2024bitsfusion}.
Additional details are provided in the appendix.

\noindent\textbf{Models and datasets.} 
We evaluate our method on both unconditional and conditional image generation models. 
For unconditional generation, we use the LDM-8 \cite{rombach2022high} model trained on the LSUN~\cite{yu2015lsun} Church $256\times256$, and the LDM-4 model trained on the LSUN Bedroom $256\times256$ and FFHQ \cite{karras2019style} $256\times256$. 
For a conditional generation, we use the LDM-4 model trained on the ImageNet \cite{deng2009imagenet} $256\times256$ and the text-conditional model StableDiffusion v1.4.

\noindent\textbf{Evaluation metrics.}
We evaluate the performance of quantized models using several standard metrics to evaluate image quality, diversity, and fidelity.
For all settings, FID \cite{heusel2017gans} and sFID \cite{nash2021generating} are used to evaluate the quality and diversity of generated images. 
In conditional generation tasks, we also include IS~\cite{salimans2016improved}, LPIPS \cite{zhang2018perceptual}, SSIM, and PSNR to further evaluate the sample quality under given conditions and how closely the generated images align with those produced by the full-precision model, ensuring the generated output faithfully reproduces the intended content.
For text-conditional generation, we report CLIP score \cite{hessel2021clipscore} to quantify alignment between generated images and text prompts.

\tabCond

\vspace{-2mm}
\subsection{Main results}
\vspace{-2mm}
\noindent\textbf{Unconditional image generation.}
We generated 50K samples using a DDIM sampler \cite{song2020denoising} with an eta of 1.0.
To reflect real-world deployment scenarios \cite{yin2024one,chen2024pixart,betker2023improving,kohler2024imagine} where computational cost is critical, we used a small number of sampling steps—specifically, 20 steps.
As shown in \cref{tab:uncond}, our method consistently demonstrates superior performance across all settings and outperforms most existing methods. Notably, baseline methods which do not consider outliers suffer significant performance degradation at W4A6, whereas our approach maintains high FID and sFID.
On the LSUN Church, our method achieves comparable results but is slightly outperformed by EDA-DM \cite{liu2024enhanced} in the W4A8 setting.
However, EDA-DM exhibits inconsistent performance across other datasets. 
Since EDA-DM is orthogonal to ours, integrating their data distribution alignment with our method could potentially yield even better results. 
Overall, our method demonstrates superior performance across diverse datasets by successfully managing outliers. 

\tabTxtToimg

\noindent\textbf{Class-conditional image generation.}
We evaluate class-conditional image generation on the ImageNet dataset using 50K images, employing 20 denoising steps with eta of 0 and a classifier-free guidance (CFG) scale of 3.0.
As shown in \cref{tab:cond}, our method consistently outperforms existing methods across all settings. 
Notably, at the low bit-width configuration of W4A6, where prior methods suffer significant performance degradation, our approach maintains high performance. 
EDA-DM, which showed promising results on specific datasets in the unconditional setting, fails to generalize to the large-scale ImageNet dataset. 
Furthermore, our method achieves superior results in LPIPS, SSIM, and PSNR across all settings, indicating that our quantized model generates outputs closest to those of the full-precision model under the same conditions. 
This is further supported by qualitative comparisons in \cref{fig:visualize}.

\noindent\textbf{Text-guided image generation.}
In this experiment, we sampled text prompts from MS-COCO  \cite{lin2014microsoft} to generate 10K images at $512 \times 512$ resolution, using 50 steps with eta of 0 and a CFG scale of 7.5.
As reported in \cref{tab:t2i}, similar to the unconditional and class-conditional generation, existing methods show significant performance degradation at W4A6, whereas our method demonstrates strong performance across nearly all metrics. 
Specifically, our method achieves the best LPIPS, SSIM, PSNR, and CLIP scores, indicating that the outputs not only closely match the full-precision model in perceptual and structural similarity but also maintain strong semantic alignment with the given text. 

\vspace{-2mm}
\subsection{Ablation study}
\vspace{-2mm}

We conducted an ablation study on LDM-4 on FFHQ at W4A8 to verify each component in our proposed method.

\noindent\textbf{Analysis of the proposed method.}
\cref{tab:ablation} presents the impact of progressively adding each component of our method. 
Starting from the baseline, introducing Learned Equivalent Scaling redistributes quantization difficulty caused by outliers, resulting in substantial metric improvements.
Adding \emph{Adaptive Timestep Weighting} further enhances performance by prioritizing early denoising steps, which, despite lower quantization errors, significantly impact final quality due to error accumulation. Finally, applying PTS to skip connection layers with high inter-channel variance yields the best overall results.

\noindent\textbf{Effect of adaptive timestep weighting.}
\cref{tab:ablation_weighting} shows the effectiveness of adaptive timestep weighting. Simple heuristics, such as linearly or quadratically increasing weights for early timesteps, underperform uniform weighting.
While early steps have smaller errors but a strong impact, later steps with larger errors are equally important.
A fixed weighting scheme may overlook critical steps, as some layers exhibit small errors in later timesteps.
In contrast, our adaptive strategy dynamically adjusts to these variations, achieving superior performance.

\noindent\textbf{Effect of the voting algorithm.}
\cref{tab:ablation_pot}  demonstrates the effectiveness of our voting algorithm for selecting PTS factors. 
The MSE-based approach, which directly minimizes quantization error, tends to overfit on a calibration set, leading to suboptimal performance. 
In contrast, our voting algorithm finds robust PTS factors by aggregating statistical consensus across samples, resulting in better generalization. Moreover, applying PTS only to skip connection layers, rather than all layers, proves to be more effective due to the high inter-channel variance present specifically in these layers. This targeted application further improves performance, as evidenced by the superior metrics achieved.

\tabAblation
\tabAblationWeighting
\tabAblationPot
\vspace{-2mm}
\section{Conclusion}
\vspace{-2mm}
We present a new approach that addresses the outlier problem in diffusion model quantization.
By integrating Learned Equivalent Scaling with an Adaptive Timestep Weighting, we redistribute quantization difficulty between weights and activations and prioritize critical timesteps.
Additionally, we propose a Power-of-Two Scaling, accompanied by a robust voting algorithm, to handle layers with extreme outliers, such as skip connections.
Extensive experiments in various datasets and generation tasks demonstrate that our method consistently outperforms existing post-training quantization techniques, even under ultra-low bit-width like W4A6, where prior methods often fail.
\section*{Acknowledgements}
This work was supported by Institute of Information \& Communications Technology Planning \& Evaluation(IITP) grant funded by the Korea government(MSIT) (RS-2024-00439020, Developing Sustainable, Real-Time Generative AI for Multimodal Interaction, SW Starlab).

{
    \small
    \bibliographystyle{ieeenat_fullname}
    \bibliography{main}
}

\ifarxiv
\clearpage
\appendix
\tabHyper

\section{Implementation details}
This section provides a more detailed description of the experimental implementation presented in the main manuscript.
\cref{tab:hyper} summarizes the hyperparameters used across all experiments for W4A8. The number of sampling steps for generating calibration data is denoted by $T$, while n represents the amount of calibration data sampled per step. 
For training, we set the batch size to $B$ and use iteration to denote the number of training steps required to learn the LES factors $\tau$. The adaptive timestep weighting loss is controlled by $\alpha$, where a higher value prioritizes optimization on early denoising steps. $\kappa$ determines the agreement threshold for selecting PTS factors, ensuring stability in the voting mechanism. The momentum parameter $\xi$ smooths the moving average update of accumulated loss values, preventing abrupt changes in timestep weights. $D$ represents the range of candidate PTS factors, allowing for optimal scaling adjustments across channels.
Specifically, the calibration dataset for the unconditional generation consists of 256 randomly generated samples, sampled using a 20-step DDIM sampler. By including all samples from the intermediate steps, the total number of calibration data points is 5120. For a class-conditional generation, we further utilize calibration data with classifier-free guidance ~\cite{ho2022classifier} (cfg), where a guidance scale is set to 3.0. For text-guided generation, the data is sampled using a 25-step DDIM sampler, resulting in a total of 6400 calibration data points.
These hyperparameters are carefully tuned for different datasets and model architectures to achieve robust and stable quantization performance.

\section{Experiment setup details}
We build upon official PyTorch implementation\footnote{\url{https://github.com/CompVis/stable-diffusion}} of LDM and used provided pre-trained models.
For a fair comparison, all methods quantize each layer except the input and output layers, which remain in full precision per common practice.
Specifically, the original implementaion of PTQD \cite{he2024ptqd} and TFMQ-DM \cite{huang2024tfmq} do not quantize \textit{skip-connection}, \textit{downsample}, and \textit{upsample} layers.
Thus, we modify their code to quantize those layers for fair comparison.
Then we directly run original code provided by the baselines.

To control for metric fluctuations caused by differences in generated samples, we unified all baselines into a single codebase and ensured identical samples were generated using the same random seed. 
Evaluation was conducted using the implementation of guided-diffusion\footnote{\url{https://github.com/openai/guided-diffusion}}. 

\section{Quantization granularity}
In this section, we provide a detailed explanation of quantization granularity.
This expands on the concepts introduced in the Preliminary section of the main paper, providing additional details.
If you are already familiar with quantization, you may skip this section.
The basic quantization and dequantization functions follow Eq. 3 in the main paper. 
Note that modern libraries implement convolutions using algorithms such as img2col \cite{chellapilla2006high} or implicit GEMM \cite{chetlur2014cudnn}, making them functionally equivalent to linear layers. Therefore, we describe our approach in the context of linear layers.
To recap, the linear layer defined in the main text is given by:
\begin{equation}
    \mathbf{Y} = \mathbf{X}\mathbf{W} \approx (s^{(\mathbf{X})} \mathbf{\tilde{X}})(s^{(\mathbf{W})} \mathbf{\tilde{W}})
\end{equation}
where $\mathbf{X} \in \mathbb{R}^{B\times C_{in}}$ is the activation, $\mathbf{W} \in \mathbb{R}^{C_{in}\times C_{out}}$ is the weight, and $\mathbf{Y} \in \mathbb{R}^{B\times C_{out}}$ is the output. $s^{(\mathbf{X})}$ and $s^{(\mathbf{W})}$ denote the scales for $\mathbf{X}$ and $\mathbf{W}$, respectively.

To investigate which quantization granularity is appropriate, we first examine per-element quantization—the finest granularity—where each element of a weight or activation has its own scale. 
Matrix multiplication with per-element quantization can be expressed as:
\begin{equation}
    \mathbf{Y}_{ij} \approx \left( \sum_{k=1}^{p} (s_{ik}^{(\mathbf{X})}\mathbf{\tilde{X}}_{ik}) \cdot (s_{kj}^{(\mathbf{W})}\mathbf{\tilde{W}}_{kj}) \right) .
\end{equation}
However, since the scales are floating-point values, including them inside the summation results in floating-point matrix multiplication, defeating the purpose of quantization.
To enable efficient integer matrix multiplication, it is necessary to factor out the scale $s$ from the summation. 
This requires the scales to be independent of \( k \) \cite{wu2020integer}.
Specifically, activation scale $s^{(\mathbf{X})}$ must be constant across columns (i.e., one scale per row), and  weight scale $s^{(\mathbf{W})}$ must be constant across rows (i.e., one scale per column).
Under this condition, the matrix multiplication can be formulated as:
\begin{equation}
    \mathbf{Y}_{ij} \approx s_{i}^{(\mathbf{X})} s_{j}^{(\mathbf{W})} \left(\sum_{k=1}^{C_{in}} \mathbf{\tilde{X}}_{ik} \cdot \mathbf{\tilde{W}}_{kj} \right) .
    \label{eq:supp_matmul}
\end{equation}
This formulation shows that integer matrix multiplication is feasible when activations are quantized with per-tensor or per-sample (row) granularity and weights are quantized in per-tensor or per-channel (column) granularity.
Since per-sample quantization requires online computation, per-tensor quantization is generally used for activations.
In this paper, we use per-tensor activation quantization, meaning that $s^{(\mathbf{X})}$ is a scalar value.
However, for consistency with the general formulation, we retain the index $i$ rather than removing it.

\section{Details of quantization approach}
In this section, we provide a detailed explanation of the quantization process, combined with proposed Learned Equivalent Scaling (LES) and Power-of-Two Scaling (PTS). 
Note that LES are applied to all layers and LES combined with PTS are only applied to skip connection layers.
Before introducing each method, we first recap the basic quantization process.

\noindent\textbf{Basic quantization.}
The standard quantization and dequantization functions for activations and weights are defined as follows:
\small{
\begin{align}
    \mathbf{\tilde{X}} &= \text{clamp}\left(\left\lfloor \frac{\mathbf{X}}{s^{(\mathbf{X})}} \right\rceil , l, u\right), \mathbf{X} \approx s^{(\mathbf{X})} \cdot \mathbf{\tilde{X}}, \\
    \mathbf{\tilde{W}} &= \text{clamp}\left(\left\lfloor \frac{\mathbf{W}}{s^{(\mathbf{W})}} \right\rceil , l, u\right), \mathbf{W} \approx s^{(\mathbf{W})} \cdot \mathbf{\tilde{W}},
\end{align}
}\normalsize
where $\lfloor\cdot\rceil$ denotes round-to-nearest-integer operator.
$\text{clamp} (\cdot)$ truncates values to $[l, u]$, yielding a low-bit representation $\mathbf{\tilde{X}}$ of $\mathbf{X}$. Here, the range $[l, u]$ is determined by the bit-width of the quantization. When $s$ is a vector, $\mathbf{X} / s$ implies channel-wise division.

\noindent\textbf{Quantization with LES.}
To recap Eq. 5 from the main paper, matrix multiplication with Equivalent Scaling can be expressed as:
\begin{equation}
    \mathbf{Y} = \left(\mathbf{X} / \tau\right) \left(\tau^\top \odot \mathbf{W}\right) = \mathbf{\hat{X}} \mathbf{\hat{W}},
\end{equation}
where $/$ and $\odot$ denote channel-wise division and multiplication, respectively.
We apply quantization and dequantization to the scaled activation $\mathbf{\hat{X}}$ and weight $\mathbf{\hat{W}}$, formulated as:
\small{
\begin{align}
    \mathbf{\tilde{X}} &= \text{clamp}\left(\left\lfloor \frac{\mathbf{\hat{X}}}{s^{(\mathbf{X})}} \right\rceil , l, u\right), \mathbf{\hat{X}} \approx s^{(\mathbf{X})} \cdot \mathbf{\tilde{X}}, \\
    \mathbf{\tilde{W}} &= \text{clamp}\left(\left\lfloor \frac{\mathbf{\hat{W}}}{s^{(\mathbf{W})}} \right\rceil , l, u\right), \mathbf{\hat{W}} \approx s^{(\mathbf{W})} \cdot \mathbf{\tilde{W}}.
\end{align}
}\normalsize
However, this formulation presents a key challenge.
Since the weight $\mathbf{W}$ remains fixed, $\tau^\top \odot \mathbf{W}$ can be precomputed and stored for direct use. In contrast, activation $\mathbf{X}$ change dynamically, making it impractical to precompute and store their scaled values.
As a result, directly applying this method would require dividing by $\tau$  at every inference step just before quantization, introducing additional computational overhead. To resolve this, we leverage the following transformation, which allows $\tau$ to be seamlessly fused into the scale factor:
\begin{equation}
    \frac{\mathbf{\hat{X}}} { s^{(\mathbf{X})} } = \frac{\mathbf{X} / \tau} {  s^{(\mathbf{X})} } =\frac{\mathbf{X}} { \tau \odot s^{(\mathbf{X})} }.
\end{equation}
This eliminates the need to divide activations by $\tau$ at every step, effectively reducing the computational overhead.
This optimization is feasible because we use static quantization, where the activation scale remains constant.
In contrast, dynamic quantization, which updates the activation scale at each step, does not allow fusing $\tau$ into the scale.
The quantization functions incorporating this method are formulated as follows:
\small{
\begin{align}
    \mathbf{\tilde{X}} &= \text{clamp}\left(\left\lfloor \frac{\mathbf{X}} { \{\tau \odot s^{(\mathbf{X})} \} } \right\rceil , l, u \right) \\
    \mathbf{\tilde{W}} &= \text{clamp}\left(\left\lfloor \frac{ \{ \tau^\top \odot \mathbf{W} \} }{s^{(\mathbf{W})}} \right\rceil , l, u \right), 
\end{align}
}\normalsize
where $\{\cdot\}$ indicates pre-computed values that are stored in memory.
The matrix multiplication with dequantization remains the same as in \cref{eq:supp_matmul}.
This approach allows us to eliminate additional overhead while efficiently integrating LES into the quantization process.

\noindent\textbf{Quantization with LES and PTS.}
For skip connection layers in residual block, where extreme outliers are observed, PTS is applied alongside LES.
The quantization and dequantization functions incorporating both LES and PTS are formulated as follows:
\small{
\begin{align}
    \mathbf{\tilde{X}} &= \text{clamp}\left(\left\lfloor \frac{\mathbf{X}} { \{ 2^{\boldsymbol{\delta}} \odot \tau \odot s^{(\mathbf{X})} \} } \right\rceil , l, u \right), \mathbf{\hat{X}} \approx  (2^{\boldsymbol{\delta}} \odot s^{(\mathbf{X})}) \cdot \mathbf{\tilde{X}} \\
    \mathbf{\tilde{W}} &= \text{clamp}\left(\left\lfloor \frac{ \{ \tau^\top \odot \mathbf{W} \} }{s^{(\mathbf{W})}} \right\rceil , l, u \right), \mathbf{\hat{W}} \approx s^{(\mathbf{W})} \cdot \mathbf{\tilde{W}}.
\end{align}
}\normalsize
Additionally, the matrix multiplication with dequantization can be expressed as:
\begin{equation}
    \mathbf{Y}_{ij}  \approx  \left(\sum_{k=1}^{C_{in}} (2^{\delta_k} s_{i}^{(\mathbf{X})} \mathbf{\tilde{X}}_{ik}) \cdot (s_{j}^{(\mathbf{W})}\mathbf{\tilde{W}}_{kj}) \right).
\end{equation}
Since the PTS factor $2^{\delta_k}$ is indexed by $k$, it cannot be factored out of the summation.
Moreover, modern GPU architectures do not natively support multiply-bitshift-add operations, making direct computation of the above formulation inefficient.
To address this, we apply bit-shifting to weights immediately after loading them during kernel execution, ensuring efficient computation:
\begin{equation}
    \mathbf{\tilde{W}}^{shifted}_{kj} = 2^{\delta_k}\mathbf{\tilde{W}}_{kj} = \mathbf{\tilde{W}}_{kj} \ll \delta_k,
    \label{eq:supp_weightshift}
\end{equation}
where $\ll$ denotes the left-bit-shift operation. 
Dequantization functions incorporating \cref{eq:supp_weightshift} are formulated as:
\begin{equation}
    \\
    \mathbf{Y}_{ij}  \approx s_{i}^{(\mathbf{X})} s_{j}^{(\mathbf{W})} \left(\sum_{k=1}^{C_{in}} \mathbf{\tilde{X}}_{ik} \cdot \mathbf{\tilde{W}}^{shifted}_{kj} \right).
\end{equation}
As a result, PTS effectively handles extreme outliers with minimal computational overhead, requiring only a bit-shifting operation on weight.

\figGemmSpeed
\section{Speedup of Power-of-Two Scaling}
To evaluate the speedup effect of Power-of-Two Scaling in quantization, we implemented a custom CUDA kernel that integrates quantization, bit-shifting on weights, GEMM (General Matrix Multiplication), and dequantization.
The comparison of latency between our custom W4A8 GEMM kernel and PyTorch FP32 GEMM is shown in \cref{fig:gemm_speed}.
Our method achieves up to $5.17\times$ speedup over FP32 at M=3072, demonstrating the efficiency of our approach.
Although bit-shifting introduces minor additional overhead, these results confirm that it can be handled highly efficiently in practice.
Additionally, since Power-of-Two Scaling is applied only to a small subset of the network (specifically, skip connection layers with extreme outliers), its overall impact on network latency remains minimal.

\section{Limitation and future Work}
In this paper, we consider robust quantization to alleviate the inter-channel variance of the neural network, which is often overlooked by the current quantization methods for diffusion models.
Consequently, we propose Learned Equivalent Scaling and channel-wise Power-of-Two Scaling, the optimization processes that consider the inter-channel variance and the iterative nature of the generative process in diffusion models.

As our approach follows the common configuration used in previous research and does not contain specific assumptions for the calibration dataset or intermediate distribution of diffusion trajectory, some settings for quantization can be further optimized to improve the upper bound of the proposed method.
For instance, we have adopted uniform timestep sampling for calibration data, following Q-Diffusion~\cite{li2023q}. However, one can freely adopt other data composition strategies such as sampling from normally distributed timesteps~\cite{shang2023post} or aligning more closely with original training samples~\cite{liu2024enhanced}. Furthermore, various quantization techniques, including noise correction~\cite{he2024ptqd,yao2024timestep} and introducing timestep-specific quantization parameters~\cite{huang2024tfmq,so2024temporal}, can be orthogonally adapted to our method.

Nevertheless, various quantization methods for diffusion models, including ours, only consider reconstructing the output of diffusion models exactly with low-bit precision. 
Moving towards robustness and efficiency in more extreme low-bit quantization, such as 2- or 3-bit representation, analyzing the feature map of diffusion models and prioritizing the salient feature could prove more effective, such as utilizing mixed-precision weight~\cite{sui2024bitsfusion}.
In addition, extending the our approach to include various downstream tasks that require efficient fine-tuning, such as personalized text-to-image diffusion models~\cite{ruiz2023dreambooth,zhang2023adding}, holds significant potential for the practical application of our method.

\section{Societal impact}
While our research primarily focuses on the quantization of diffusion models to achieve efficient generation, it is important to acknowledge the broader societal impact of generative models. By enabling more efficient generation, our approach could make generative models more accessible to a wider range of users, increasing the potential for both positive and negative applications such as generating deepfakes, NSFW content, or copyright-infringing materials. 
Integrating safeguards like erasing NSFW content from diffusion models~\cite{gandikota2023erasing} into the quantization process should be considered to mitigate potential risks.

\section{More Visual Results}
\cref{fig:vis_imagenet,fig:vis_bedroom,fig:vis_church,fig:vis_ffhq} shows samples generated with W4A6 quantized models using state-of-the art methods on various datasets. 
The results demonstrate that the proposed quantization method effectively generates high-quality images while maintaining the structure of samples from full-precision models.

\figVisBedroom
\figVisChurch
\figVisFFHQ
\figVisImagenet
\fi

\end{document}